\documentclass[runningheads]{llncs}

\usepackage[final,year=2026,ID=6788]{eccv}

\usepackage{eccvabbrv}

\usepackage{graphicx}
\usepackage{booktabs}

\usepackage[accsupp]{axessibility}  

\usepackage{hyperref}

\usepackage{orcidlink}

\usepackage{multirow}
\usepackage{booktabs}
\usepackage{graphicx} 

\usepackage[table]{xcolor}  
\definecolor{best}{RGB}{255,230,230}
\newcommand{\best}[1]{\cellcolor{best}\textbf{#1}}

\begin{document}

\title{NeuIDO: Neural Intrinsic Dynamics Operator for Physics-Informed 4D World Models} 

\titlerunning{NeuIDO}

\author{Jiajing Lin \inst{1}\orcidlink{0009-0004-6004-5398} \and Xin Zhang \inst{2}\orcidlink{0009-0000-3601-5296} \and
Jianhua Sun\thanks{Corresponding author: Jianhua Sun (gothic@sjtu.edu.cn).}\inst{1}\orcidlink{0000-0002-1030-5575}}

\authorrunning{J. Lin et al.}

\institute{
School of Artificial Intelligence, Shanghai Jiao Tong University, China \and
School of Informatics, Xiamen University, China
}

\maketitle

\begin{abstract}
World models aim to capture environmental dynamics and predict future trajectories, showing growing potential for embodied intelligence. 
Physics-informed 4D generation integrates physical simulation to predict 3D object interactions, offering a promising pathway toward world models. However, this paradigm relies on manually imposed dynamical assumptions rather than internalizing world dynamics, and thus still leaves a gap toward a true world model.
To bridge this gap, we propose NeuIDO, a novel world dynamics modeling framework that learns a unified intrinsic dynamics representation from visual observations, advancing physics-informed 4D generation toward a world model. 
Specifically, we formulate world modeling as a neural operator learning problem and introduce a two-stage training strategy to learn a generalizable mapping from the visual observation distribution to the intrinsic dynamics distribution.
Building on this observation-dynamics mapping, NeuIDO enables zero-shot dynamics inference directly from videos and can be further aligned with complex real-world dynamics via few-shot adaptation. 
Extensive experiments demonstrate that NeuIDO effectively unifies the intrinsic dynamics underlying diverse visual observations into a shared representation and rapidly infers dynamics in novel scenes.
\end{abstract}

\begin{figure*}[!t]
	\centering
	\includegraphics[width=1\linewidth]{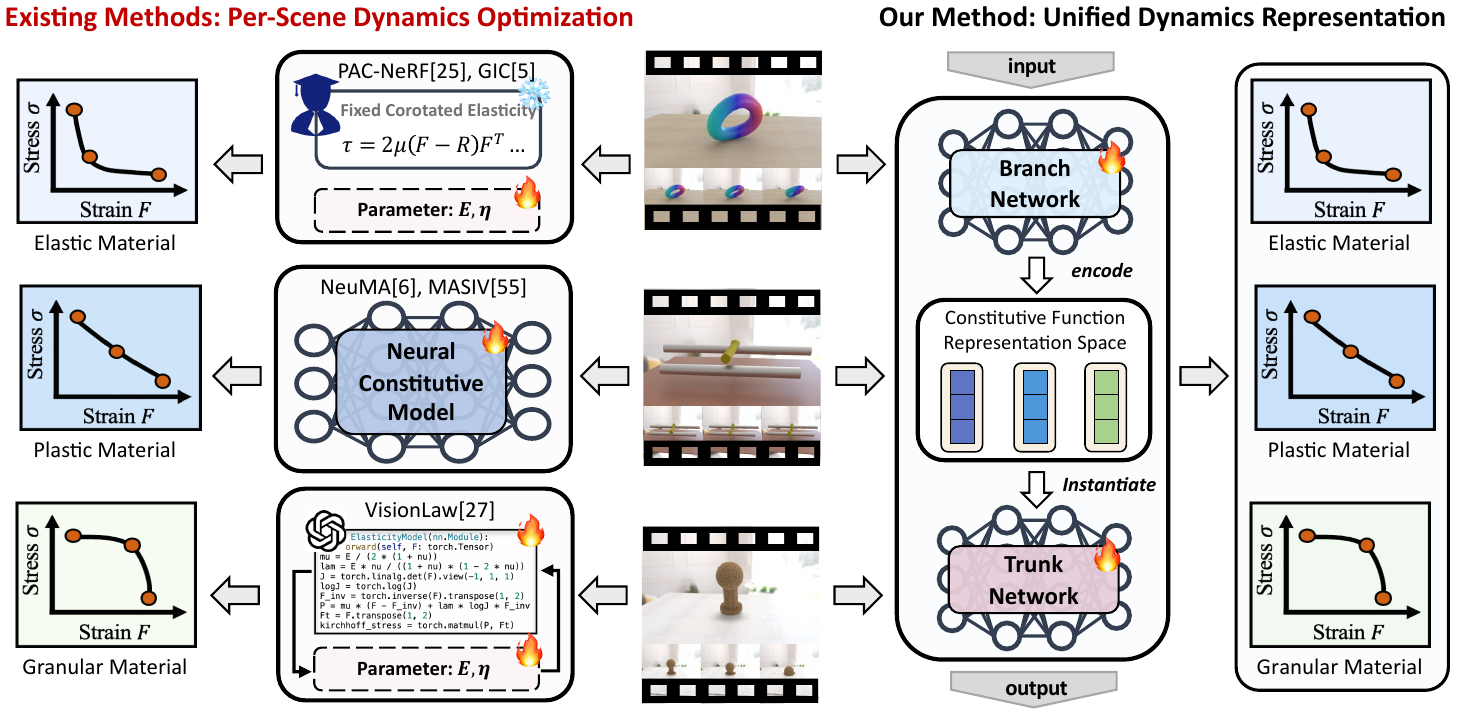}
	\caption{
    Existing approaches typically perform per-scene dynamics optimization, which is computationally expensive and generalizes poorly to unseen scenes.
    In contrast, {\em NeuIDO} learns a unified intrinsic dynamics representation that enables efficient zero-shot generalization, 
    moving physics-informed 4D generation toward world modeling.
    }
    \label{fig:intro}
\end{figure*}

\section{Introduction}
World models~\cite{worldmodel, JEPA, Sora} provide agents with an “imagination space” for planning and decision-making, and have shown great potential in applications such as autonomous driving~\cite{GAIA-1, GAIA-2} and robotics~\cite{embodiedsurvey, sun2025digital}.
Most current world models~\cite{DreamV1, DreamV2, V-JEPA} learn environment dynamics from large-scale 2D data (e.g., images or videos) and predict future states via 2D frame generation.
Despite progress under this purely data-driven paradigm, the absence of explicit physical constraints and limited 3D spatial reasoning (e.g., inaccurate object layouts and relative relations) often leads to predictions that violate fundamental physical principles~\cite{howfarlaw} (e.g. conservation laws), 
limiting reliability for embodied decision-making.

In contrast, physics-informed 4D generation~\cite{Physgaussian, Phys4DGen} explicitly incorporates physical simulators to evolve 3D scenes over time. Such a paradigm enables physically consistent movement prediction and provides an explicit 3D spatiotemporal representation, thus offering a promising avenue to mitigate the limitations of purely data-driven world models. 
However, physical simulation relies on access to the system’s intrinsic dynamics, whose core lies in the constitutive law~\cite{continuum} with specified material parameters that govern the object’s response under applied forces. These laws are typically imposed through hand-crafted assumptions.

Humans can intuitively infer an object’s intrinsic dynamics through vision and predict its future evolution.
To achieve such intuitive physical inference, recent methods have attempted to integrate differentiable renderers, such as Neural Radiance Fields (NeRF) and 3D Gaussian Splatting (3DGS)~\cite{NeRF,3DGS}, with differentiable simulators~\cite{MPM} in an end-to-end paradigm. Specifically, some methods~\cite{PAC-NeRF, GIC, PhysDreamer} estimate unknown material parameters through visual supervision. 
However, they rely on hand-crafted constitutive laws, making it difficult to align with the complex dynamics of the real world. Furthermore, another class of methods~\cite{NeuMA, MASIV, VisionLaw} attempts to learn neural or symbolic constitutive models from visual observations. 
However, these approaches independently infer object intrinsic dynamics from scratch for each scene, preventing the learning of a shared cross-scene dynamics structure and limiting efficient generalization to unseen scenes. 
As a result, they remain short of a true world model.

To overcome the above challenges, we propose {\em NeuIDO}, a novel world dynamics modeling framework that learns a unified intrinsic dynamics representation from multi-scene video observations and supports efficient intrinsic dynamics inference for unseen scenes—thereby advancing physics-informed 4D generation toward a world model.
Inspired by the Universal Approximation Theorem (UAT) for neural operators—which states that neural operators can approximate continuous mappings between infinite-dimensional function spaces—we cast world dynamics modeling as an operator learning problem. Specifically, we learn a mapping from an input (e.g., videos) function space to an output (e.g., constitutive laws) function space, enabling diverse intrinsic dynamics across scenes to be modeled within a unified framework.
To reduce the complexity of operator learning, we further propose a two-stage training strategy: (i) pretraining the structural representation space of the constitutive function from large-scale known laws to inject dynamical priors; and (ii) aligning videos with this learned representation space so that each observation is interpreted as a manifestation of intrinsic dynamics.
Leveraging this observation–dynamics operator mapping, {\em NeuIDO} enables real-time zero-shot intrinsic dynamics inference from videos. Furthermore, to better capture complex real-world dynamics, it supports efficient few-shot adaptation via lightweight fine-tuning on visual observations.
Our contributions are summarized as follows:
\begin{itemize}
    \item We propose a world dynamics modeling framework that learns a unified intrinsic dynamics representation from visual observations across multiple scenes, advancing physics-informed 4D generation toward a true world model.
    \item We formulate world dynamics modeling as an operator learning problem and address optimization challenges with a two-stage strategy: (i) structured dynamics representation learning and (ii) observation-to-dynamics alignment.
    \item Extensive experiments on synthetic and real-world datasets demonstrate that {\em NeuIDO} effectively models shared cross-scene dynamics structure and achieves real-time zero-shot generalization with efficient few-shot adaptation.
\end{itemize}

\section{Related Work}
\subsection{Physics-Informed 4D Interaction}
In recent years, 3D representations~\cite{NeRF, 3DGS, wu2025textsplat} have increasingly been coupled with physics simulators~\cite{MPM, PBD, XPBD} to enable interactive 4D dynamics~\cite{Phys4DGen, Phy124, WonderPlay}.
PIE-NeRF~\cite{PIE-NeRF} couples NeRF with a mesh-free elastodynamics simulator, PhysGaussian~\cite{Physgaussian} combines 3DGS with an MPM simulator~\cite{MPM}, and VR-GS~\cite{VR-GS} enables real-time VR interaction through fast XPBD-based simulation~\cite{XPBD}.
However, these approaches rely on expert-specified constitutive laws and carefully tuned parameters.
PAC-NeRF~\cite{PAC-NeRF} and GIC~\cite{GIC} perform differentiable simulation-based identification from multi-view videos. Another line of work, such as PhysDreamer, DreamPhysics, Physics3D, and PhysFlow~\cite{PhysDreamer, DreamPhysics, Physics3D, PhysFlow}, distills dynamic priors from video diffusion models to guide estimation. Along this direction, OmniPhysGS~\cite{OmniPhysGS} further assigns each Gaussian kernel a constitutive model from a predefined set. Nevertheless, these methods still rely on expert-defined constitutive laws, limiting their ability to capture complex real-world dynamics. To reduce this dependency, NeuMA~\cite{NeuMA} and MASIV~\cite{MASIV} optimize neural constitutive models to align simulations with visual observations, though the learned models are implicit and lack interpretability. VisionLaw~\cite{VisionLaw} further advances this direction by discovering explicit constitutive-law expressions through bilevel optimization.
Despite these advances, existing methods typically learn a separate intrinsic dynamics for each scene. 
\subsection{Data-Driven World Model}
World models~\cite{daydreamer,rssm1,worldmodel} aim to understand and predict environment dynamics, enabling applications such as autonomous driving~\cite{Drivedreamerv1, Drivedreamerv2}, robotics~\cite{Robodreamer,robotwd1, wei2026physically, sun2025discovering, sun2025arti}, and VR/AR~\cite{sorasurvey}. 
Existing approaches broadly fall into three families: SSM-based, JEPA-based, and Transformer-based methods. World Models~\cite{worldmodel} learns compact latent representations and RNN-based latent dynamics from high-dimensional observations, serving as an early precursor to later RSSM-based world models. Dreamer family~\cite{DreamV1, DreamV2, DreamV3} extends this paradigm to model-based reinforcement learning via latent imagination. 
JEPA~\cite{JEPA} learns predictive world representations by predicting embeddings of masked regions rather than reconstructing pixels. I-JEPA~\cite{I-JEPA} and V-JEPA~\cite{V-JEPA} apply this idea to images and videos, respectively, through context-to-masked-block prediction without hand-crafted augmentations.
Transformer-based world models~\cite{Transdreamer, transformerwd1,transformerwd2} replace recurrent dynamics with attention-based sequence modeling~\cite{Transformer} to better capture long-horizon dependencies. More recently, large-scale generative world simulators such as Sora~\cite{Sora} and Genie~\cite{Genie} learn controllable environment dynamics from video, enabling long-horizon generation, while 3D-aware generative world models~\cite{3Dwdsurvey} (e.g., Genie 2, Explorer, LWMs) further couple generative modeling with neural scene representations to construct consistent 3D worlds.
Despite their effectiveness, existing world models largely follow a purely data-driven paradigm and lack explicit physical constraints, which often leads to physically inconsistent predictions. In this work, we instead explicitly model world dynamics, moving toward a physics-informed 4D world model.

\section{Methodology}
\subsection{Problem Setup and Overview}
\begin{figure*}[!t]
	\centering
	\includegraphics[width=1\linewidth]{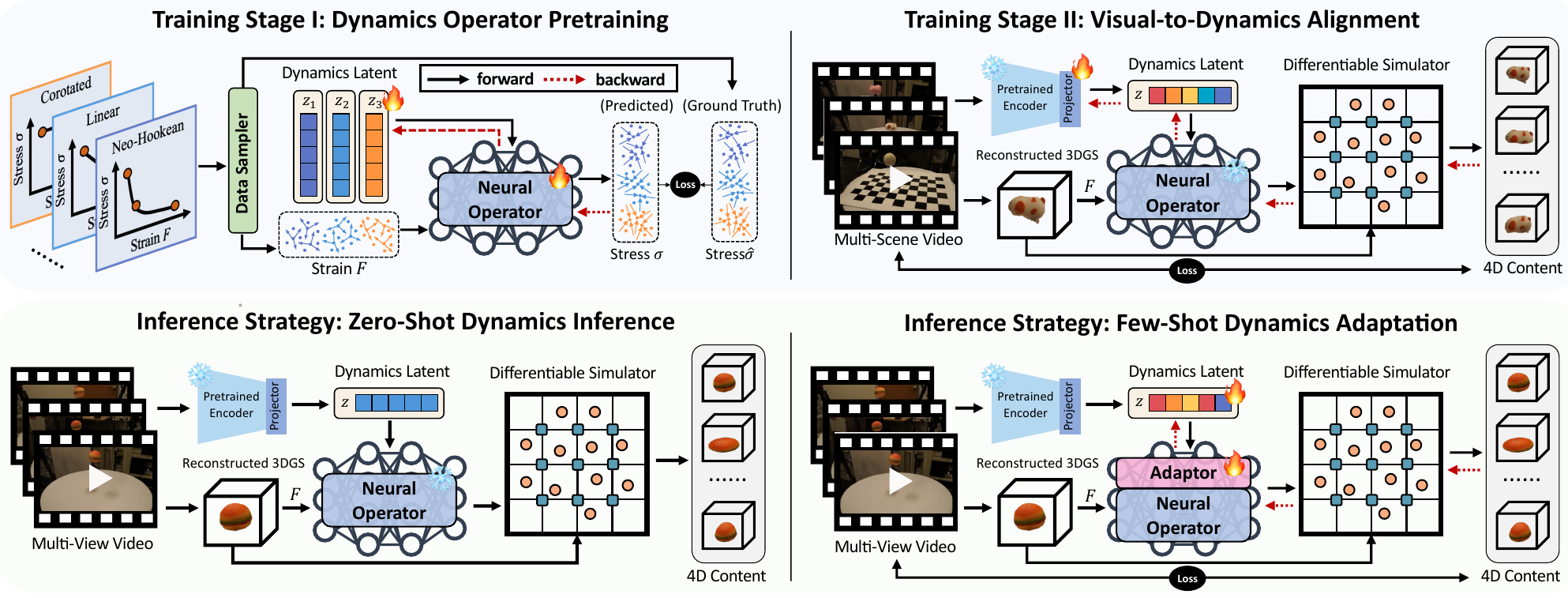}
	\caption{\textbf{Pipeline of NeuIDO.} In Stage I, we pretrain the operator on a collection of known constitutive laws to learn a structured representation space of constitutive functions, injecting rich dynamics priors. In Stage II, we train a projector to align visual observations with this representation space, establishing a connection between videos and intrinsic dynamics. Under the zero-shot paradigm, the operator immediately infers plausible intrinsic dynamics for unseen scenes. Under the few-shot paradigm, lightweight adaptation further aligns the operator to out-of-distribution dynamics.}
    \label{fig:pipeline}
\end{figure*}

In this paper, we aim to learn a unified representation that captures the diverse intrinsic dynamics underlying visual observations across multiple scenes, while enabling efficient dynamics inference in novel scenarios. 
Formally, let $K_V \subset \mathbb{R}^{d_v}$ denote the domain of video functions, and let $\mathcal{C}(K_V)$ be the corresponding space of continuous video functions.
In practice, we consider a compact subset $\mathcal{V} \subset \mathcal{C}(K_V)$ representing the set of valid video observations that sufficiently capture object motions governed by scene-specific dynamics.
We consider a set of $N$ dynamic scenes, each providing calibrated multi-view videos $\mathbf{u} \in \mathcal{V}$.
Let $\mathcal{C}(K_C)$ be the constitutive function space defined on the constitutive-input domain $K_C \subset \mathbb{R}^{d_c}$.
Our goal is to learn an operator:
\begin{equation}
    \widehat{\mathcal{G}}: \mathcal{V} \to \mathcal{C}(K_C)
\end{equation}
which maps the video function space to the constitutive function space, such that diverse scene dynamics are embedded into a unified representation.
In Sec.~\ref{method:framework}, we formulate a neural intrinsic dynamics operator.
In Sec.~\ref{method:learning}, we propose a two-stage training strategy to learn the unified dynamics representation.
In Sec.~\ref{method:inference}, we present zero-shot and few-shot inference strategies for generalization to novel scenes.
In addition, we provide a theoretical analysis in the Appendix to support the feasibility of our method for modeling world dynamics.
The overall architecture of {\em NeuIDO} is illustrated in Fig.~\ref{fig:pipeline}.

\subsection{Unified Intrinsic Dynamics Representation Framework}
\label{method:framework}
\subsubsection{Intrinsic Dynamics Operator Parameterization.}
We parameterize the mapping $\widehat{\mathcal{G}}$ from observation to dynamics using a neural intrinsic dynamics operator.
Formally, $\widehat{\mathcal{G}}$ acts on an input video function $\mathbf{u}$ to output a continuous constitutive function, $\widehat{\mathcal{G}}(\mathbf{u})$. This function can then be evaluated at any given constitutive input $\boldsymbol{\xi} \in K_C$ (e.g., deformation gradient $\mathbf{F}$) to yield the specific physical response, $\widehat{\mathcal{G}}(\mathbf{u})(\boldsymbol{\xi})$. Specifically, inspired by DeepONet~\cite{DeepONet}, {\em NeuIDO} is formulated as a branch–trunk operator architecture.
The branch network uses a dynamics encoder $\mathbf{E}$ to extract $\mathbf{z} \in \mathbb{R}^{d}$ from the input video, such that $\mathbf{z} = \mathbf{E}(\mathbf{u})$, capturing the intrinsic dynamics governing the observed motion.
This latent representation is then mapped to the coefficient vector of the operator expansion through a multilayer perceptron (MLP), producing $\boldsymbol{\lambda} \in \mathbb{R}^{D}$.
The trunk network parameterizes shared basis functions $\{\phi_k(\cdot)\}_{k=1}^D$, and evaluates them at $\boldsymbol{\xi}$ to output the vector $\{\phi_k(\boldsymbol{\xi})\}_{k=1}^D$.
The video-dependent constitutive function $\widehat{\mathcal{G}}(\mathbf{u})$ is then represented as a linear combination of these basis functions weighted by the coefficients.
\begin{equation}
\label{eq:operator}
\widehat{\mathcal{G}}(\mathbf{u})(\boldsymbol{\xi})=\sum_{k=1}^{D} \lambda_k\,\phi_k(\boldsymbol{\xi}).
\end{equation}
Under this formulation, the trunk network learns a shared basis spanning the constitutive function space, while the branch network predicts scene-specific coefficients that instantiate the constitutive relation for each observation.
This design enables {\em NeuIDO} to effectively fit the mapping between visual observations and intrinsic dynamics, while exploiting the generalization of neural operators to enable efficient inference in unseen scenarios.

\subsubsection{Trunk Network: Intrinsic Dynamics Basis Representation.}
The trunk network parameterizes a shared set of constitutive basis functions $\{\phi_k(\cdot)\}_{k=1}^D$ across different scenes. For a query point $\boldsymbol{\xi} \in K_C$, we evaluate these basis functions at $\xi$ to obtain their basis values $\{\phi_k(\boldsymbol{\xi})\}_{k=1}^D$, which are then linearly combined with the branch-predicted coefficients in Eq.~\ref{eq:operator}.
In differentiable MPM, two types of constitutive relations must be defined: an elastic component that describes the reversible material response under deformation, and a plastic component that captures irreversible evolution beyond the elastic limit.
Accordingly, we parameterize two trunk networks to match these two constitutive mechanisms. 

{(i) Elastic trunk network.}
The elastic trunk takes the deformation gradient $\mathbf{F}$ as input and outputs stress basis values
$\{\boldsymbol{\phi}_k^E(\mathbf{F})\}_{k=1}^{D}$.
Given on the visual observation $\mathbf{u}$, the elastic constitutive function $\widehat{\mathcal{G}}^E(\mathbf{u})$ is represented as:
\begin{align}
\widehat{\mathcal{G}}^E(\mathbf{u})(\mathbf{F})
&= \sum_{k=1}^{D} \lambda_k^E\, \phi_k^E(\mathbf{F}).
\end{align}
The coefficient vector $\boldsymbol{\lambda}^E$ is obtained by mapping the dynamics latent through the elastic MLP head of the branch network.

{(ii) Plastic trunk network.}
Similarly, the plastic trunk takes $\mathbf{F}$ as input and outputs corrected deformation-gradient basis values
$\{\mathbf{\phi}_k^P(\mathbf{F})\}_{k=1}^{D}$.
The scene-specific plastic constitutive function $\widehat{\mathcal{G}}^P(\mathbf{u})$ is parameterized in a residual form:
\begin{align}
\widehat{\mathcal{G}}^P(\mathbf{u})(\mathbf{F})
&= \mathbf{F} + \sum_{k=1}^{D} \lambda_k^P\, \phi_k^P(\mathbf{F}).
\end{align}
The coefficient vector $\boldsymbol{\lambda}^P$ is obtained by mapping the same dynamics latent through the plastic MLP head of the branch network.

\subsubsection{Branch Network: Intrinsic Dynamics Latent Identification.}
The branch network aims to extract scene-dependent dynamics features from visual observations and transform them into coefficient vectors that, together with the trunk network, instantiate the constitutive operators.
To this end, we introduce a dynamics encoder $\mathbf{E}$ that maps an input video $\mathbf{u}\in\mathbb{R}^{L\times H\times W\times 3}$ to a compact dynamics latent $\mathbf{z} \in \mathbb{R}^{d}$, which captures the underlying dynamics governing the observed motion.
Concretely, $\mathbf{E}$ consists of three stages:
(i) we compute the optical flow $\mathbf{u}_f\in\mathbb{R}^{(L-1)\times H\times W\times 3}$ between consecutive frames to emphasize motion cues;
(ii) we employ the pretrained optical flow encoder from Tora~\cite{Tora} to embed the flow fields into a spatiotemporal feature representation
$\mathbf{z}^0\in\mathbb{R}^{\frac{L-1}{4}\times\frac{H}{8}\times\frac{W}{8}\times 3}$, which captures motion patterns across frames;
(iii) a projector network maps $\mathbf{z}^0$ into the compact dynamics latent $\mathbf{z}$.
The dynamics latent $\mathbf{z}$ is then mapped by two MLP heads to produce elastic and plastic coefficient vectors,
$\boldsymbol{\lambda}^E\in\mathbb{R}^{D}$ and $\boldsymbol{\lambda}^P\in\mathbb{R}^{D}$, respectively, which are used to weight the corresponding trunk-evaluated bases.

\subsection{Unified Intrinsic Dynamics Representation Learning}
\label{method:learning}
Due to video supervision being sparse and indirect with respect to intrinsic dynamics, learning the observation-to-dynamics operator in an end-to-end manner is challenging. 
In contrast, numerous known constitutive relations provide strong supervision covering diverse dynamical mechanisms, offering powerful dynamics priors.
To exploit this advantage, we adopt a two-stage training strategy. In Stage I, the operator is pretrained on known constitutive relations to learn a structured representation space of constitutive functions. In Stage II, video observations are aligned with this pretrained space, enabling the operator to infer intrinsic dynamics consistent with the observed motions.
This two-stage design effectively transfers knowledge from strongly supervised constitutive data to weakly supervised visual observations, facilitating stable learning of the mapping from visual observations to intrinsic dynamics.

\subsubsection{Stage I: Dynamics Operator Pretraining.}
This stage aims to learn a structured \emph{representation space of constitutive functions} under strong physical supervision, so that diverse intrinsic dynamics can be expressed.
This representation space is parameterized by a shared trunk basis and indexed by a learnable dynamics latent.
To construct the training data, we collect a large set of known constitutive relations and use the MPM simulator to generate material trajectories for each constitutive function.
For each trajectory $i$, we record deformation gradients and their corresponding constitutive responses over particles and timesteps. A total of $M$ trajectories are collected to form the training dataset:
\begin{equation}
\mathcal{D}=\Big\{\big(\mathbf{F}^{(i)}_{p,t},\,\mathbf{S}^{(i)}_{p,t},\,\widehat{\mathbf{F}}^{(i)}_{p,t}\big)\Big\}, \quad p=1,\dots,P,\ t=1,\dots,T ,    
\end{equation}
where $\mathbf{F}$ is the deformation gradient, $\mathbf{S}$ is the elastic stress, and $\hat{\mathbf{F}}$ denotes the plasticity-induced corrected deformation gradient.
As visual observations are not yet introduced in this stage,  we bypass the dynamics encoder $\mathbf{E}$ and instead assign each trajectory a learnable dynamics latent $\mathbf{z}^{(i)}$. 
This latent serves as an index to instantiate a specific constitutive function within the shared trunk basis.
Concretely, we map $\mathbf{z}^{(i)}$ to coefficient vectors $\boldsymbol{\mathbf{\lambda}}^{E,(i)}$ and $\boldsymbol{\mathbf{\lambda}}^{P,(i)}$ via lightweight elastic and plastic MLP heads, and evaluate the trunk networks on $\mathbf{F}$ to form latent-conditioned constitutive operators.
The operator parameters (excluding the dynamics encoder) and the dynamics latents are jointly optimized by minimizing reconstruction errors of elastic and plastic responses:
\begin{equation}
\min_{\theta^E,\theta^P,\{\mathbf{z}^{(i)}\}}
\sum_{i=1}^{M}\sum_{p,t}
\Big(
\mathcal{L}_r\big(\widehat{\mathcal{G}}^{E}(\mathbf{z}^{(i)})(\mathbf{F}^{(i)}_{p,t}),\,\mathbf{S}^{(i)}_{p,t}\big)
+
\mathcal{L}_r\big(\widehat{\mathcal{G}}^{P}(\mathbf{z}^{(i)})(\mathbf{F}^{(i)}_{p,t}),\,\widehat{\mathbf{F}}^{(i)}_{p,t}\big)
\Big),
\end{equation}
where $\mathcal{L}_r$ denotes the relative loss~\cite{neuraloperatorlibrary}, and $\theta^E$ and $\theta^P$ represent the parameters of the elastic and plastic trunk networks, respectively. 
After Stage I, the model captures shared structures across constitutive functions and learns a physically grounded latent space for material dynamics.
This pretrained space serves as a strong prior for Stage II, where videos are aligned to it by learning the mapping from visual observations to dynamics latents via the branch video encoder $\mathbf{E}$.

\subsubsection{Stage II: Visual-to-Dynamics Alignment.}
Stage II aligns visual observations with the constitutive representation space learned in Stage I, establishing a connection between videos and intrinsic dynamics.
Specifically, we freeze the trunk networks and optimize only the branch-side dynamics encoder $\mathbf{E}$, which maps a video observation $\mathbf{u}$ to a dynamics latent $\mathbf{z} = \mathbf{E}(\mathbf{u})$.
Given a training set of $N$ dynamic scenes, where each scene provides multi-view videos capturing object motions governed by intrinsic dynamics, each scene $i$ is treated as one training instance.
For each scene, the predicted dynamics latent $\mathbf{z}^{(i)}$ instantiates elastic constitutive function $\widehat{\mathcal{G}}^E(\mathbf{z}^{(i)})$ and plastic constitutive function $\widehat{\mathcal{G}}^P(\mathbf{z}^{(i)})$ within the frozen operator space.
These functions are then used to drive a differentiable physics-integrated 4D simulation pipeline $\mathcal{R}$ (see Appendix for details) to generate rendered videos
$\widetilde{V}^{(i)} = \mathcal{R}(\widehat{\mathcal{G}}^E(\mathbf{z}^{(i)}), \widehat{\mathcal{G}}^P(\mathbf{z}^{(i)}))$.
We supervise the dynamics encoder $\mathbf{E}$ by minimizing the reconstruction error between rendered and observed videos over all scenes:
\begin{equation}
\min_{\omega}\;
\sum_{i=1}^{N}
\mathcal{L}_{\text{mse}}\Big(\mathcal{R}\big(\widehat{\mathcal{G}}^E(\mathbf{z}^{(i)}), \widehat{\mathcal{G}}^P(\mathbf{z}^{(i)})\big),
\mathbf{u}^{(i)}
\Big),
\end{equation}
where $\omega$ denotes the parameters of the dynamics encoder $\mathbf{E}$.
By optimizing this objective over diverse scenes, the dynamics encoder learns a generalizable mapping from visual motion patterns to the intrinsic-dynamics latents that index the pretrained constitutive representation space.

\subsection{Inference with Unified Intrinsic Dynamics Representation}
\label{method:inference}
\subsubsection{Zero-Shot Dynamics Inference.}
After the two-stage training, {\em NeuIDO} has learned (i) a pretrained constitutive representation space and (ii) a shared dynamics encoder that maps visual observations to the dynamics latents indexing this space. Therefore, {\em NeuIDO} can infer intrinsic dynamics for an unseen scene in a purely feed-forward manner. Given an input video observation $\mathbf{u}$ from a new scene, we first extract its dynamics latent using the branch encoder $\mathbf{z} = \mathbf{E}(\mathbf{u})$. The latent $\mathbf{z}$ then instantiates a scene-specific constitutive function within the pretrained operator space, denoted as $\widehat{\mathcal{G}}^E(\mathbf{z})$ and $\widehat{\mathcal{G}}^P(\mathbf{z})$. We plug these constitutive functions into the physics-driven simulation pipeline to roll out the 4D dynamics. Importantly, zero-shot inference requires \emph{no additional optimization}. This enables efficient constitutive inference and interaction simulation in previously unseen scenes, reflecting the learned cross-scene generalization of the visual-to-dynamics mapping.

\subsubsection{Few-Shot Dynamics Adaptation.}
While the zero-shot paradigm enables immediate dynamics identification in unseen scenes, it may encounter challenges in complex real-world scenarios where the target dynamics significantly deviate from the training distribution.
To improve alignment in such cases, we perform a few-shot adaptation mechanism.
Instead of updating the entire model, we insert a lightweight adaptor into the trunk network and jointly refine the scene-specific dynamics latent $\mathbf{z}$ predicted by the dynamics encoder $\mathbf{E}$.
This design enables efficient adaptation while preventing overfitting.
Given visual observations $\mathbf{u}$ from a scene, we jointly optimize the adaptor parameters $\theta_{\text{adp}}^E$, $\theta_\text{adp}^P$ and latent $\mathbf{z}$ by minimizing the reconstruction error:
\begin{equation}
\min_{\theta_{\text{adp}}^E, \theta_\text{adp}^P, \mathbf{z}}
\mathcal{L}_\text{mse}
\Big(
\mathcal{R}\big({\widehat{\mathcal{G}}}^E(\mathbf{z};\theta_{\text{adp}}^E), {\widehat{\mathcal{G}}}^P(\mathbf{z};\theta_{\text{adp}}^P)\big),
\mathbf{u}
\Big),
\end{equation}
where $\widehat{\mathcal{G}}(\mathbf{z};\theta_{\text{adp}})$ denotes the constitutive operator instantiated by latent $z$ with the trunk adaptor applied.
This lightweight adaptation enables the model to rapidly specialize to out-of-distribution dynamical environments.

\section{Experiments}
\subsection{Experimental Setup}
\subsubsection{Implementation Details.}
We follow a DeepONet-style operator~\cite{DeepONet} parameterization with a branch–trunk architecture, and set the dynamics latent dimension to 128 (see the Appendix for architectural details). During Training Stage I, each material trajectory is assigned a learnable dynamics latent initialized from $\mathcal{N}(0,1)$. These latents are jointly optimized together with the trunk network for 10 epochs using a batch size of $10^4$. In Training Stage II, each visual scene is treated as one training instance. We freeze the operator and optimize only the projector in the branch network. Training alternates over all $N$ scenes for a total of  $N \times 100$ epochs. For few-shot adaptation, we insert LoRA modules~\cite{LoRA} into the trunk ($r = 16$, $\alpha = 16$) and jointly optimize the adaptor parameters and the scene-specific latent for 100 epochs. All dynamics rollouts are performed with an MPM simulator in a $[0,1]^3$  domain under gravity ($9.8 m/s^2$). We use grid resolutions of $32^3$ for synthetic data and $70^3$ for real-world data. For each scene, we initialize the simulation by reconstructing a 3DGS representation from the first frame following NeuMA~\cite{NeuMA}, and supervise alignment using 2-view videos for synthetic scenes and 3-view videos for real scenes. All optimization is performed using the AdamW optimizer with a cosine learning rate scheduler. All experiments are conducted on a single NVIDIA H100 (80GB) GPU.

\begin{table*}[t]
\centering
\setlength{\tabcolsep}{7pt}      
\renewcommand{\arraystretch}{1.15}
\resizebox{\linewidth}{!}
{
    \begin{tabular}{@{} c l | c c c c c c c @{\hskip\tabcolsep}}
    \toprule
    & \textbf{Method} & \textbf{Ball} & \textbf{Cat} & \textbf{Bottle} & \textbf{Duck} & \textbf{Pawn} & \textbf{Fish} & \textbf{Average} \\
    \midrule
    \multirow{4}{*}{\rotatebox{90}{CD$\downarrow$}}
     & PAC-NeRF~\cite{PAC-NeRF} & 516.300 & 15.380 & 2.210 & 137.730 & 15.470 & 1.71 & 114.80 \\
     & NCLaw~\cite{NCLaw} & 56.690 & 2.350 & 0.920 & 11.970 & 3.910 & 1.300 & 12.860 \\
     & NeuMA~\cite{NeuMA} & 1.271 & 0.679 & 0.680 & 3.506 & 0.882 & 0.953 & 1.329 \\
     & VisionLaw~\cite{VisionLaw} & \best{1.085} & 0.768 & \best{0.639} & 5.205 & 0.942 & 1.078 & 1.620 \\
    \midrule
     & Ours(w/o finetune) & 1.159 & 0.844 & 0.714 & 5.113 & 0.973 & 1.141 & 1.643 \\
     & Ours(w/ finetune) & 1.356 & \best{0.617} & 0.687 & \best{2.850} & \best{0.848} & \best{0.937} & \best{1.216} \\
    \midrule
    
    \end{tabular}
}
\caption{\textbf{Comparison of chamfer distance on the synthetic dataset.} Lower chamfer distance indicates better consistency with ground-truth dynamics.}
\label{tab:synthetic datatset}

\end{table*}

\subsubsection{Datasets and Metrics.}
For Stage I operator pretraining, we construct a constitutive function dataset covering three classical material families: elastic, plasticine, and sand. For visual dynamics modeling, we adopt six dynamic scenes from NeuMA~\cite{NeuMA} as synthetic data. Each scene provides RGB videos from 10 viewpoints with 400 frames per view, along with ground-truth particle trajectories. For real-world evaluation, we use three scenes from Spring-Gaus~\cite{Spring-Gaus}. Each scene provides RGB videos from three viewpoints with 19 frames per view. To further evaluate zero-shot generalization, we additionally collect 10 scenes with distinct elastic dynamics from PAC-NeRF~\cite{PAC-NeRF}, using eight for training and two for testing. Additional dataset details are reported in the Appendix. To evaluate the consistency between the modeled dynamics and the ground-truth dynamics, we (1) use chamfer distance to measure the distribution discrepancy between simulated and ground-truth particle trajectories, and (2) report PSNR, SSIM, and LPIPS to assess visual reconstruction quality.

\subsection{Modeling Observed Intrinsic Dynamics}
\label{sec: 4.2}
\subsubsection{Comparison on Synthetic Datasets.}
\begin{figure*}[!t]
	\centering
	\includegraphics[width=1\linewidth]{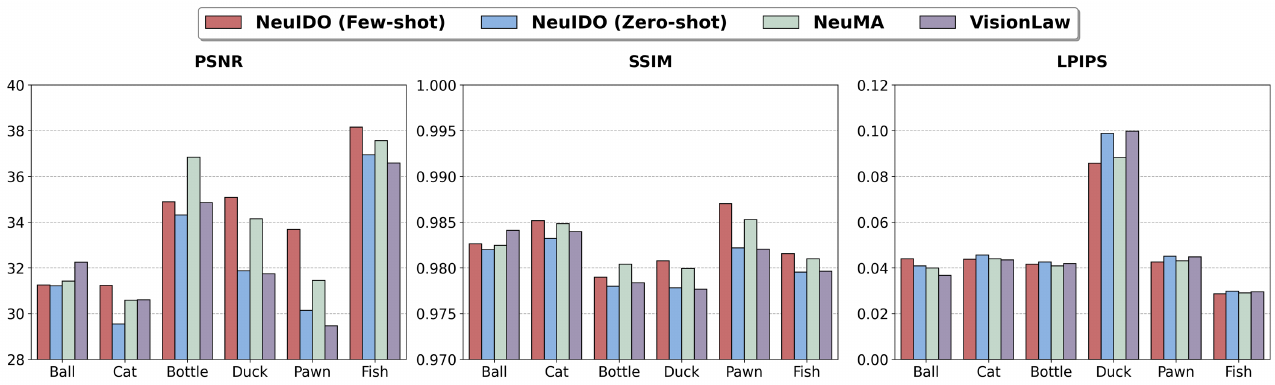}
	\caption{\textbf{Comparison of visual fidelity on the synthetic dataset}. We report average PSNR, SSIM and LPIPS between simulated and ground-truth videos on novel views. Higher PSNR/SSIM and lower LPIPS indicate better reconstruction quality.}
    \label{fig:synthetic dataset}
\end{figure*}

To evaluate intrinsic dynamics modeling within the training distribution, we conduct quantitative comparisons on the synthetic dataset. The chamfer distance (CD) results are reported in Tab.~\ref{tab:synthetic datatset}.
Under the zero-shot setting, {\em NeuIDO} achieves performance comparable to state-of-the-art methods. Meanwhile, unlike PAC-NeRF, which relies on predefined constitutive models, NCLaw, which requires particle trajectory supervision, and NeuMA and VisionLaw, which perform scene-specific modeling, {\em NeuIDO} learns a unified dynamics operator from multi-scene visual observations. We observe that zero-shot inference may exhibit larger errors in scenes with significant distribution shifts from the pretraining data (e.g., JellyDuck). 
With few-shot adaptation, the average CD is reduced from 1.643 to 1.216, outperforming all baselines.
Notably, {\em NeuIDO} reaches this performance within only 100 adaptation iterations, while NeuMA requires around 1000 iterations even when initialized from an NCLaw-pretrained model, demonstrating its ability to efficiently adapt to complex scenes.
Fig.~\ref{fig:synthetic dataset} further reports PSNR, SSIM, and LPIPS on unseen viewpoints. 
These trends align with the CD results in Tab.~\ref{tab:synthetic datatset}. Notably, although VisionLaw achieves the lowest CD on the Bottle scene, it does not yield the best visual quality, suggesting that accurate dynamics modeling does not necessarily translate to superior visual reconstruction. 

\begin{figure*}[!t]
	\centering
	\includegraphics[width=1\linewidth]{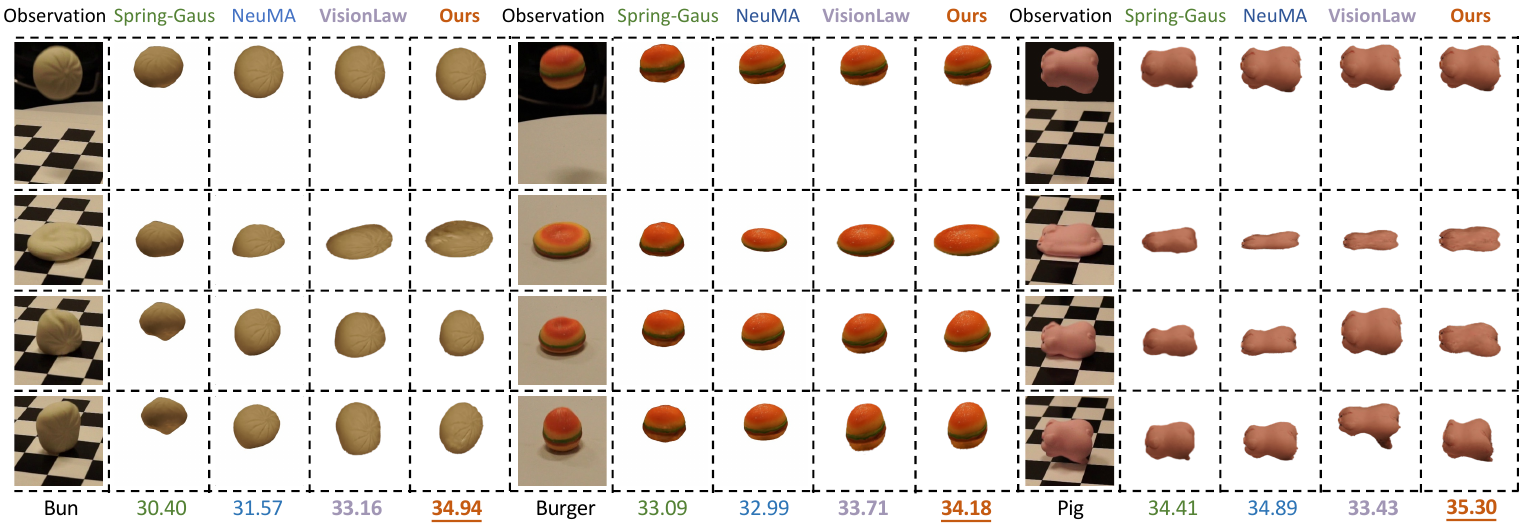}
	\caption{\textbf{Comparison on real-world datasets.} Quantitative metrics (i.e., PSNR) between the predicted and observed frames are reported in the bottom row. Our method achieves the best visual dynamics reconstruction.}
    \label{fig:real-world dataset}
\end{figure*}

\subsubsection{Comparison on Real-World Datasets.}
To evaluate real-world dynamics modeling, Fig.~\ref{fig:real-world dataset} compares {\em NeuIDO} (with few-shot adaptation) with Spring-Gaus, NeuMA, and VisionLaw on real-world datasets. Spring-Gaus embeds a spring–mass system into 3DGS for elastic modeling, but its limited dynamic expressivity struggles with complex real-world dynamics. NeuMA integrates MPM simulation with neural networks and outperforms Spring-Gaus, though its performance heavily relies on an NCLaw-pretrained model. VisionLaw introduces physical inductive biases via LLMs to guide constitutive evolution, yet its optimization is prone to local minima in complex real-world scenes. As shown in Fig.~\ref{fig:real-world dataset}, {\em NeuIDO} yields reconstructions closer to real observations and achieves the highest PSNR across all real scenes. We attribute this improvement to the unified dynamics representation learned by {\em NeuIDO}, which provides strong physical priors for few-shot adaptation, thus enabling more effective alignment to complex real-world dynamics. Together, these results demonstrate that {\em NeuIDO} effectively captures intrinsic dynamics in real-world scenarios, providing a practical pathway toward physics-informed world models and a stronger foundation for dynamics understanding in embodied agents interacting with real environments.

\begin{figure*}[!t]
	\centering
	\includegraphics[width=1\linewidth]{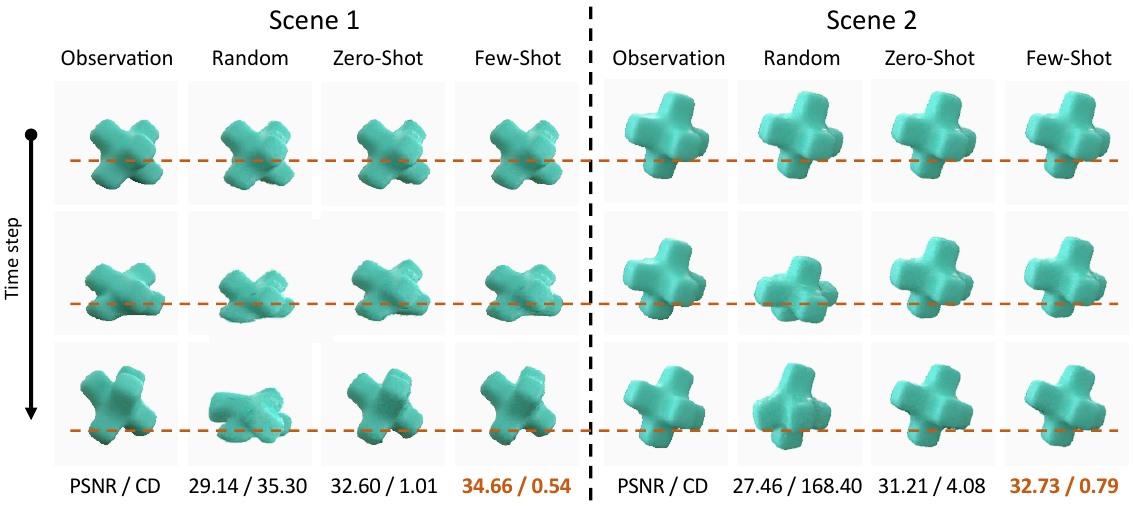}
	\caption{\textbf{Generalization on unseen scenes.} Quantitative metrics (PSNR $\uparrow$ / CD $\downarrow$) and qualitative results for two test scenes. The random baseline uses initialized dynamics latents randomly. {\em NeuIDO} produces dynamics close to the ground-truth observation in the zero-shot setting and further improves with few-shot adaptation.
    }
    \label{fig:unseen scene generalization}
\end{figure*}

\begin{figure*}[!t]
	\centering
	\includegraphics[width=1\linewidth]{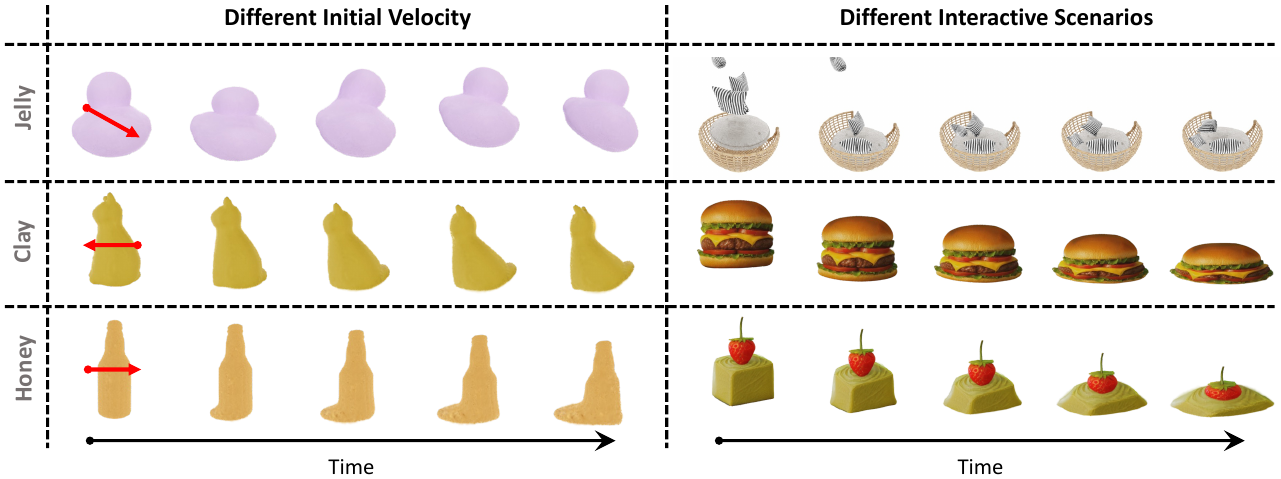}
	\caption{\textbf{Transfer of modeled intrinsic dynamics to new simulation conditions.} The left text indicates the intrinsic dynamics learned by {\em NeuIDO}. Left: Examples with different initial velocities (red arrows denote velocity directions). Right: Interactions with new simulation objects.}
    \label{fig:simulation conditions generalization}
\end{figure*}

\begin{figure*}[!t]
	\centering
	\includegraphics[width=1\linewidth]{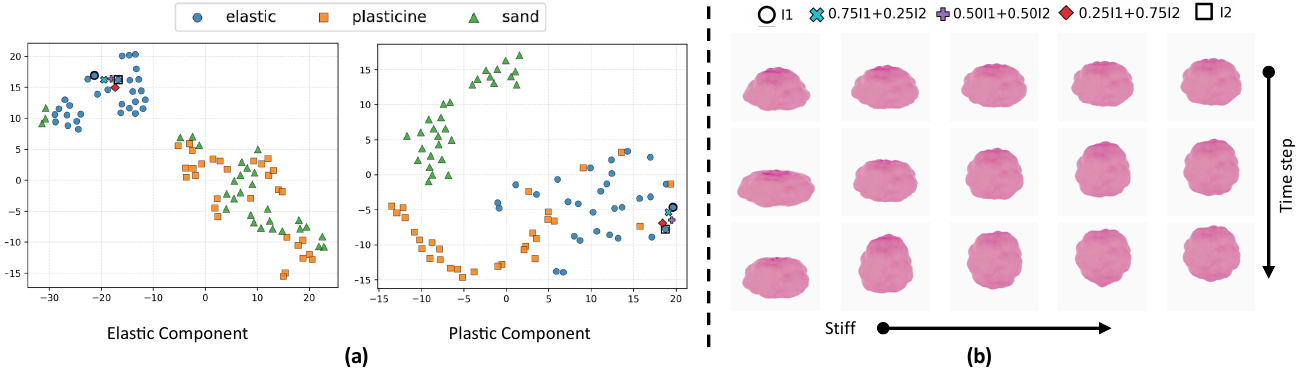}
	\caption{\textbf{Representation analysis.} (a) t-SNE visualization of Stage-I pretrained dynamics latents of 90 material trajectories.
    (b) 4D rollouts generated by interpolating between two dynamics latents.
    Top markers indicate the interpolation settings, showing a smooth transition from soft to stiff behavior.
    }
    \label{fig:representation_analysis}
\end{figure*}

\subsection{Generalization Analysis}
\subsubsection{Generalization to Unseen Scenarios.}
\label{sec: 4.3.1}
While the previous experiments evaluate intrinsic dynamics modeling within the training distribution, we further examine whether {\em NeuIDO} can infer dynamics in completely unseen scenes. Following PAC-NeRF, we select 10 scenes with diverse elastic behaviors, using eight for training and two as unseen test scenes. As shown in Fig.~\ref{fig:unseen scene generalization}, {\em NeuIDO} achieves strong zero-shot performance on the unseen scenes, significantly outperforming the random baseline in both qualitative results and quantitative metrics (PSNR / CD). Importantly, {\em NeuIDO} correctly distinguishes different physical behaviors purely from visual observations without any optimization. For example, Scene~1 exhibits softer deformation patterns, while Scene~2 shows a noticeably more rigid response. These results suggest that the model does not simply memorize the appearance of training scenes, but instead learns a transferable mapping from visual observations to intrinsic dynamics. Overall, {\em NeuIDO} generalizes beyond the training distribution and enables immediate dynamics inference in unseen environments, providing a promising foundation for physics-informed world modeling to support embodied decision-making in novel environments.

\subsubsection{Generalization to Different Simulation Conditions.}
We further evaluate whether the learned intrinsic dynamics can generalize across different simulation conditions. To this end, we transfer the modeled dynamics to 4D interaction tasks following the interaction paradigms of PhysGaussian and Phy124, while modifying the simulation setup, including the initial velocity and interacting objects. As shown in Fig.~\ref{fig:simulation conditions generalization}, the resulting dynamics remain consistent with the original observations despite changes in simulation conditions. For example, when an initial velocity toward the lower-right direction is applied, JellyDuck still exhibits clear elastic deformation and recovery behavior. Similarly, when clay dynamics are applied to Hamburger, the object reproduces the slow settling behavior characteristic of clay materials. 
These results suggest that {\em NeuIDO} learns scene-independent intrinsic dynamics that generalize across simulation conditions.

\subsection{Representation Analysis}
To examine whether {\em NeuIDO} learns a structured intrinsic dynamics representation, we visualize the Stage-I pretrained dynamics latents of 90 material trajectories using t-SNE and analyze their interpolation behavior. 
As shown in Fig.~\ref{fig:representation_analysis} (a), the latents from the same constitutive family are clustered together, while different constitutive families are clearly separated.
We further evaluate the continuity of this representation by interpolating between two dynamics latents and rolling out the corresponding 4D dynamics. 
Fig.~\ref{fig:representation_analysis} (b) reveals a smooth transition from soft to stiff motion along the interpolation.
These results show that the learned latent space is discriminative across constitutive families and continuous within the dynamics manifold, supporting {\em NeuIDO}'s zero-shot inference.

\begin{table*}[t]
\centering
\setlength{\tabcolsep}{7pt}      
\renewcommand{\arraystretch}{1.5}
\resizebox{\linewidth}{!}
{
    \begin{tabular}{@{} c c | cc | ccccccc @{}}
    \toprule
    
    & \multirow{2}{*}{\textbf{Dim}} 
    & \multicolumn{2}{c|}{\textbf{Pretraining}}
    & \multicolumn{7}{c}{\textbf{Alignment}} \\
    
    \cmidrule(lr){3-4}
    \cmidrule(lr){5-11}
    & 
    & \textbf{Elasticity} 
    & \textbf{Plasticity} 
    & \textbf{Ball} 
    & \textbf{Cat} 
    & \textbf{Bottle} 
    & \textbf{Duck} 
    & \textbf{Pawn} 
    & \textbf{Fish} 
    & \textbf{Average}\\
    
    \midrule
    
    \multirow{3}{*}{} 
    & 32  & 1.023e3  & 3.598e-8 & \best{3.721\%}  & -33.850\% & -12.497\% & -0.052\% & -2.108\% & -13.289\% &-9.679\% 
    \\
    & 64  & 8.392e2 & \best{3.211e-8} & -7.361\%  & -29.942\% & -9.340\% & \best{-0.046\%} & \best{3.240\%} & -14.247\% & -9.616\% 
    \\
    & 256 & \best{6.885e2} & 3.341e-8  & -3.229\%  & \best{4.631\%} & \best{6.397\%} & -1.573\% & -0.890\% & \best{3.418\%} & \best{1.459\%} 
    \\
    \bottomrule
    \end{tabular}
}
\caption{\textbf{Analysis of impact of dynamics latent dimensionality on operator expressiveness.} We report the MSE loss in Stage I (pretraining) and the relative change in chamfer distance with respect to the 128-dimensional in Stage II (alignment).}
\label{tab:latent dim ablation}
\end{table*}

\begin{figure*}[!t]
	\centering
	\includegraphics[width=1\linewidth]{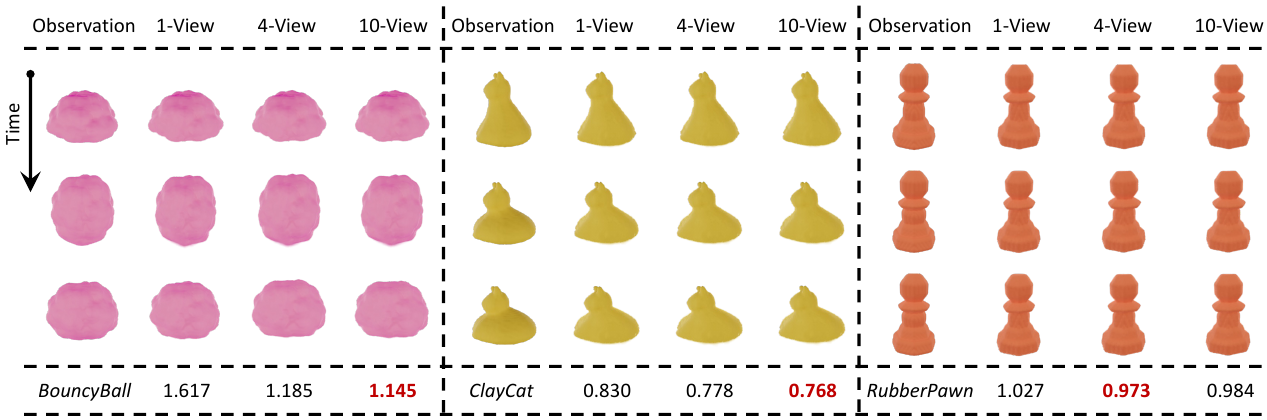}
	\caption{\textbf{Effect of multi-view supervision.} Comparisons using 1, 4, and 10 input views during Stage II alignment. Chamfer distances are shown in the bottom row.}
    \label{fig:multi view ablation}
\end{figure*}

\subsection{Ablation Study}
\subsubsection{Effect of Latent Dimensionality.}
We study the impact of dynamics latent dimensionality on operator expressiveness by evaluating different latent sizes. Specifically, we report the MSE loss in Stage I (pretraining) and relative change in chamfer distance with respect to the 128-dimensional default in Stage II (alignment). As shown in Tab.~\ref{tab:latent dim ablation}, in the pretraining phase, smaller latent dimensions (e.g., 32) result in significantly higher fitting errors, indicating insufficient capacity to index diverse constitutive functions. Increasing the latent dimension to 64 and 256 substantially reduces the error. A similar trend is observed in the alignment phase. The 32-dimensional setting degrades performance across several scenes, whereas the 256-dimensional latent achieves the best or near-best results in most cases. These results suggest that latent dimensionality directly affects the expressive capacity of the operator: higher-dimensional latents allow the model to capture more complex dynamics and improve fitting accuracy. 

\subsubsection{Effect of Multi-View Supervision.}
We analyze the impact of multi-view supervision by training the operator with 1, 4, and 10 input views during Stage II alignment (Fig.~\ref{fig:multi view ablation}). Increasing supervision from a single view to multiple views significantly improves dynamics reconstruction, as additional viewpoints help reduce geometric ambiguity and incomplete observations. However, increasing the number of views from 4 to 10 provides only marginal improvement, indicating diminishing returns from additional visual supervision. A similar trend is observed in the few-shot multi-view ablation reported in the Appendix. These results suggest that simply increasing the number of views does not continuously improve intrinsic dynamics modeling. Since intrinsic dynamics are inferred indirectly through image-space reconstruction, rendering errors may accumulate. Therefore, further improvements in dynamics reconstruction likely require stronger physical constraints rather than additional visual observations.

\section{Conclusion}
In this paper, we introduce {\em NeuIDO}, a world dynamics modeling framework that learns a unified representation of intrinsic dynamics from visual observations across scenes. By formulating world modeling as a neural operator learning problem, {\em NeuIDO} establishes a mapping from the observation space to the constitutive function space, enabling diverse scene dynamics to be captured within a shared representation. To make this feasible, we introduce a two-stage training strategy: (i) pretraining a structured representation space of constitutive functions from a collection of known constitutive relations, and (ii) aligning visual observations with this learned space to establish the observation–dynamics link. Experiments on both synthetic and real-world datasets demonstrate that {\em NeuIDO} effectively models cross-scene dynamics, enabling real-time zero-shot dynamics inference and efficient few-shot adaptation in unseen scenes. 
These results highlight the potential of {\em NeuIDO} as a step toward physics-informed world models for embodied planning and decision-making.

\section*{Acknowledgements}
This work was supported by Fundamental and Interdisciplinary Disciplines Breakthrough Plan of the Ministry of Education of China, the Shanghai Committee of Science and Technology, China (Grant No.24511103200), Shanghai Artificial Intelligence Laboratory, XPLORER PRIZE grants. 
We sincerely thank Jikuang Zhang for his valuable contributions to this work.

%

\newpage
\bibliographystyle{splncs04}
\bibliography{main}

\newpage
\appendix
\onecolumn
\section*{Appendix}


\section{Overview}
In this appendix, we will provide: i) more experimental details; ii) more experimental results; iii) physics-informed 4D generation pipeline; iv) limitation and future work; v) universal approximation theorem of the neural intrinsic dynamics operator.

\section{More Experimental Details}
\subsection{Dataset Details}
For Stage I operator pretraining, we construct a constitutive function dataset covering three classical material families: elastic, plasticine, and sand (their constitutive formulations are provided in~\ref{Classical Material Models}). For each material, we randomly sample 30 sets of material parameters, resulting in 30 distinct material instances. For each instance, we run an MPM simulation under a unified physical setup in which a cubic object evolves under gravity. Each simulation produces one trajectory that records the temporal evolution of stress and deformation gradient. In total, the dataset contains $3 \times 30$ trajectories.
For visual dynamics modeling, we adopt six dynamic scenes from NeuMA~\cite{NeuMA} as synthetic data. The dataset covers diverse material types ranging from elastic solids to granular materials and multiple object geometries. Each scene provides RGB videos from 10 viewpoints with 400 frames per view, along with ground-truth particle trajectories. To reduce computational overhead, we temporally downsample the videos by retaining one frame out of every five, resulting in 80 frames per view for training.
For real-world evaluation, we collect three real-world scenes from Spring-Gaus~\cite{Spring-Gaus}. Each scene provides RGB videos from three viewpoints with 19 frames per view. To evaluate zero-shot dynamics inference on unseen scenes, we further collect 10 elastic material scenes with distinct dynamics from the PAC-NeRF~\cite{PAC-NeRF}. Each scene contains videos from 10 viewpoints with 14 frames per view. We use eight scenes for training and two scenes for testing to assess zero-shot generalization.
In all experiments, the initial velocity is assumed to be known.

\subsection{Classical Material Models}
\label{Classical Material Models}
In this work, we use three classical materials, \emph{elastic}, \emph{plasticine}, and \emph{sand}, to generate the Stage-I training dataset. 
Following the standard constitutive formulation within MPM framework, the behavior of each material is determined by an elastic constitutive law together with a plasticity constitutive law.
\subsubsection{Elastic Material}
The dynamics of the elastic material are defined by \emph{Fixed Corotated Elasticity} together with \emph{Identity Plasticity}, i.e., purely elastic deformation without permanent plastic flow.

\paragraph{Fixed Corotated Elasticity.}
The Kirchhoff stress is defined as
\begin{equation}
\boldsymbol{\tau} = 2\mu \left( \mathbf{F} - \mathbf{R} \right)\mathbf{F}^T + \lambda J \left( J - 1 \right) \mathbf{I},
\end{equation}
where
\begin{equation}
\mathbf{F} = \mathbf{U}\boldsymbol{\Sigma}\mathbf{V}^T, \qquad \mathbf{R} = \mathbf{U}\mathbf{V}^T,
\end{equation}
are given by the singular value decomposition of the deformation gradient $\mathbf{F}$, and $J=\det(\mathbf{F})$.

\paragraph{Identity Plasticity.}
The corrected deformation gradient is simply
\begin{equation}
\mathbf{F}^{\text{corrected}} = \mathbf{F}.
\end{equation}

\subsubsection{Plasticine Material}
The dynamics of plasticine are modeled by \emph{Sigma Elasticity} together with \emph{Von Mises Plasticity}. 
This combination captures elastic response under small deformation and irreversible plastic flow once the yield criterion is exceeded.

\paragraph{Sigma Elasticity.}
The Kirchhoff stress is defined as
\begin{equation}
\label{eq: sigma elasticity}
\boldsymbol{\tau} =
\mathbf{U}
\,
\mathrm{diag}\!\left(
2\mu \log(\boldsymbol{\sigma}) +
\lambda \, \mathrm{tr}\!\left(\log(\boldsymbol{\sigma})\right)\mathbf{1}
\right)
\mathbf{U}^T,
\end{equation}
where
\begin{equation}
\mathbf{F} = \mathbf{U}\boldsymbol{\Sigma}\mathbf{V}^T,
\qquad
\boldsymbol{\Sigma} = \mathrm{diag}(\boldsymbol{\sigma}).
\end{equation}

\paragraph{Von Mises Plasticity.}
The corrected deformation gradient is defined as
\begin{equation}
\mathbf{F}^{\text{corrected}} = \mathbf{U}\,\mathcal{Z}(\boldsymbol{\Sigma})\,\mathbf{V}^T,
\end{equation}
where
\begin{equation}
\mathcal{Z}(\boldsymbol{\Sigma}) =
\begin{cases}
\boldsymbol{\Sigma}, 
& \delta\gamma \leq 0, \\[4pt]
\exp\!\left( \boldsymbol{\epsilon} - \delta\gamma 
\frac{\hat{\boldsymbol{\epsilon}}}{\|\hat{\boldsymbol{\epsilon}}\|} \right), 
& \text{otherwise},
\end{cases}
\end{equation}
and
\begin{equation}
\boldsymbol{\epsilon} = \log(\boldsymbol{\Sigma}), 
\qquad
\hat{\boldsymbol{\epsilon}} = \mathrm{dev}(\boldsymbol{\epsilon}),
\qquad
\delta\gamma = \|\hat{\boldsymbol{\epsilon}}\| - \frac{\tau_Y}{2\mu}.
\end{equation}
Here, $\tau_Y$ denotes the yield stress. 
Von Mises plasticity is widely used for materials exhibiting isotropic yielding, such as clay-like or metal-like plastic behavior.

\subsubsection{Sand Material}
The dynamics of sand are modeled by \emph{Sigma Elasticity} together with \emph{Drucker--Prager Plasticity}. 
Compared with Von Mises plasticity, the Drucker--Prager model additionally accounts for pressure-dependent yielding, which is important for granular materials such as sand.

\paragraph{Sigma Elasticity.} 
Sand material adopts the same elastic constitutive law as the plasticine material described in Eq.~\ref{eq: sigma elasticity}.

\paragraph{Drucker--Prager Plasticity.}
Given
\begin{equation}
\mathbf{F} = \mathbf{U} \boldsymbol{\Sigma} \mathbf{V}^T,
\qquad
\boldsymbol{\epsilon} = \log(\boldsymbol{\Sigma}),
\qquad
\hat{\boldsymbol{\epsilon}} = \mathrm{dev}(\boldsymbol{\epsilon}),
\end{equation}
the corrected deformation gradient is defined as
\begin{equation}
\mathbf{F}^{\text{corrected}} = \mathbf{U} \, \mathcal{Z}(\boldsymbol{\Sigma}) \, \mathbf{V}^T,
\end{equation}
with
\begin{equation}
\mathcal{Z}(\boldsymbol{\Sigma}) =
\begin{cases}
\mathbf{I}, & \text{if } \mathrm{tr}(\boldsymbol{\epsilon}) > 0, \\[4pt]
\boldsymbol{\Sigma}, & \text{if } \delta \gamma \leq 0 \ \text{and} \ \mathrm{tr}(\boldsymbol{\epsilon}) \leq 0, \\[4pt]
\exp\!\left( \boldsymbol{\epsilon} - \delta \gamma \frac{\hat{\boldsymbol{\epsilon}}}{\|\hat{\boldsymbol{\epsilon}}\|} \right), & \text{otherwise},
\end{cases}
\end{equation}
where
\begin{equation}
\delta \gamma = \|\hat{\boldsymbol{\epsilon}}\| + \alpha \frac{(d\lambda + 2\mu)\,\mathrm{tr}(\boldsymbol{\epsilon})}{2\mu},
\qquad
\alpha = \sqrt{\frac{2}{3}} \cdot \frac{2\sin\phi_f}{3 - \sin\phi_f}.
\end{equation}
Here, $\phi_f$ is the friction angle. 

\begin{figure*}[!t]
	\centering
	\includegraphics[width=1\linewidth]{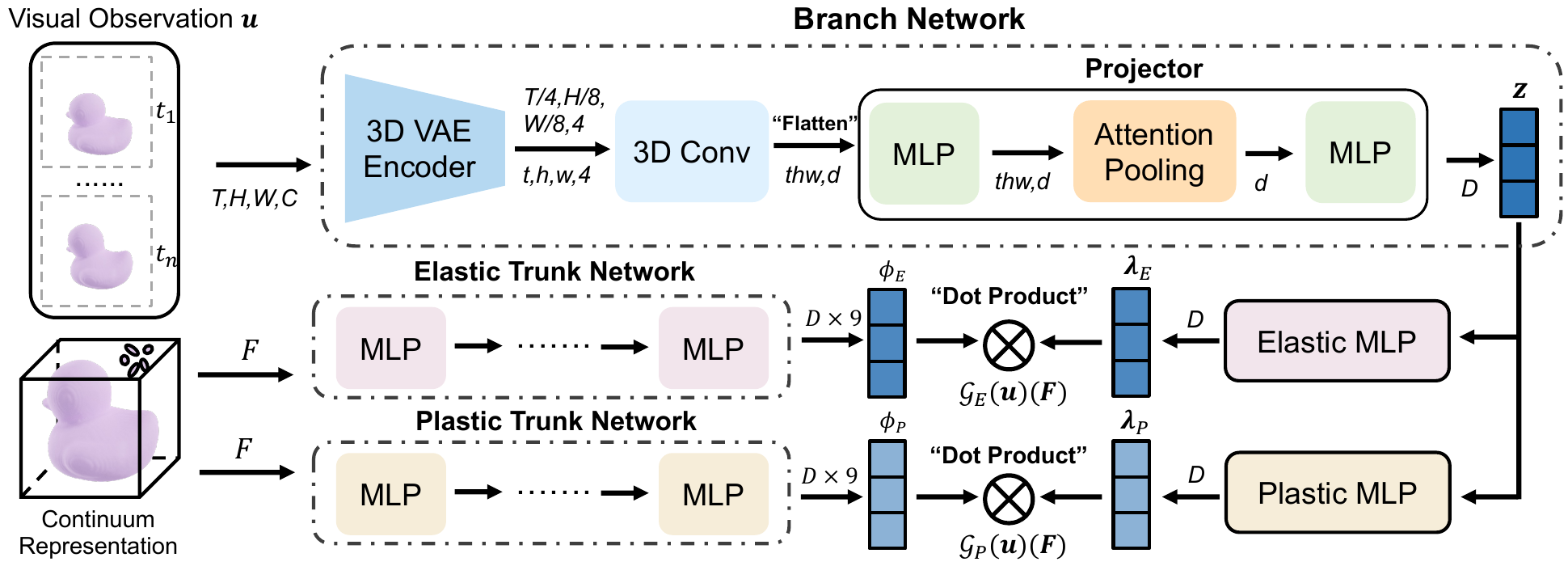}
	\caption{\textbf{Model Architecture.} The branch network encodes an input video into a dynamics latent $z$ using a 3D VAE encoder, 3D convolution, and an attention-based projector.  The latent is mapped to coefficient vectors for elastic and plastic components. Two trunk networks take the deformation gradient $F$ as input and output basis values. The final constitutive responses are obtained through dot-product interactions between the branch coefficients and trunk basis values.}
    \label{fig:model}
\end{figure*}

\subsection{Model Architecture Details}
We provide additional implementation details of the proposed {\em NeuIDO} architecture. 
As illustrated in Fig.~\ref{fig:model}, the architecture follows a branch--trunk neural operator design, where the branch network extracts coefficient vectors from visual observations and the trunk networks parameterize constitutive basis functions.

\textbf{Branch Network.}
The branch network maps an input video observation $\mathbf{u} \in \mathbb{R}^{T \times H \times W \times C}$ to a compact dynamics latent $\mathbf{z} \in \mathbb{R}^{D}$. 
The video is first processed by a pretrained 3D VAE encoder from Tora~\cite{Tora} to obtain a spatiotemporal feature volume. 
A 3D convolution layer is then applied to further aggregate motion patterns. 
The resulting features are flattened along the spatiotemporal dimensions and passed through a projector composed of an MLP, an attention pooling module, and a final MLP. 
The attention pooling layer aggregates spatiotemporal tokens into a global feature representation, which is then projected to the dynamics latent $\mathbf{z}$. 
To instantiate scene-specific constitutive functions, the latent $\mathbf{z}$ is further mapped by two independent MLP heads to produce the coefficient vectors $\boldsymbol{\lambda}_E \in \mathbb{R}^D$ and $\boldsymbol{\lambda}_P \in \mathbb{R}^D$ corresponding to the elastic and plastic components, respectively.

\textbf{Trunk Networks.}
{\em NeuIDO} employs two trunk networks with identical architectures but separate parameters for elastic and plastic modeling. 
Each trunk is implemented as a three-layer MLP that maps the deformation gradient $\mathbf{F}$ to a $D \times 9$-dimensional set of basis values, denoted as $\phi_E(\mathbf{F})$ and $\phi_P(\mathbf{F})$. 
Each basis output is $9$-dimensional because the constitutive responses (stress tensor or corrected deformation gradient) are represented as $3\times3$ matrices, which are vectorized into a $9$-dimensional form.

\textbf{Operator Instantiation.}
The final constitutive functions are obtained by dot-product interactions between the branch-predicted coefficients and the trunk-evaluated basis values. 
Specifically, the elastic response is computed from $\boldsymbol{\lambda}_E$ and $\phi_E(\mathbf{F})$, while the plastic response is computed from $\boldsymbol{\lambda}_P$ and $\phi_P(\mathbf{F})$. 
In this formulation, the branch network produces scene-dependent coefficients, whereas the trunk networks provide a shared basis representation of constitutive functions.

\subsection{Baselines}
{\em NeuIDO} is the first method to learn a unified representation of intrinsic dynamics directly from visual observations. Consequently, we conduct a best-effort comparison with several representative approaches that address related problems from different perspectives. 
All baselines are implemented following their original settings. 
\begin{itemize}
    \item PAC-NeRF~\cite{PAC-NeRF} estimates material parameters from visual observations. However, its reliance on manually specified constitutive models limits its ability to capture complex real-world dynamics.
    \item NCLaw~\cite{NCLaw} learns neural constitutive models from pre-defined particle trajectory rather than visual observations and is typically used as a neural constitutive prior.
    \item NeuMA~\cite{NeuMA} extends NCLaw by introducing a neural adaptor that aligns neural constitutive models with visual observations. Although this enables visual grounding of dynamics, NeuMA performs scene-specific modeling and does not learn a unified intrinsic dynamics representation across scenes. Among existing approaches, NeuMA is therefore the closest to our work.
    \item VisionLaw~\cite{VisionLaw} leverages large language models to introduce strong physical inductive biases for guiding the search of constitutive expressions, effectively alleviating NeuMA’s overfitting under 1-view observations. However, its performance heavily relies on the LLM-driven search for constitutive structures, which is complex and stochastic rather than driven by visual-supervision optimization. Consequently, in our 2-view experiments, VisionLaw achieves performance similar to its single-view setting.
    \item Spring-Gaus~\cite{Spring-Gaus} integrates a mass–spring system into a 3D Gaussian Splatting representation to model elastic dynamics. However, its expressiveness is limited to relatively simple elastic behaviors, making it difficult to capture complex real-world dynamics. In this work, we compare with Spring-Gaus on real-world datasets.
\end{itemize}

\begin{figure*}[!t]
	\centering
	\includegraphics[width=1\linewidth]{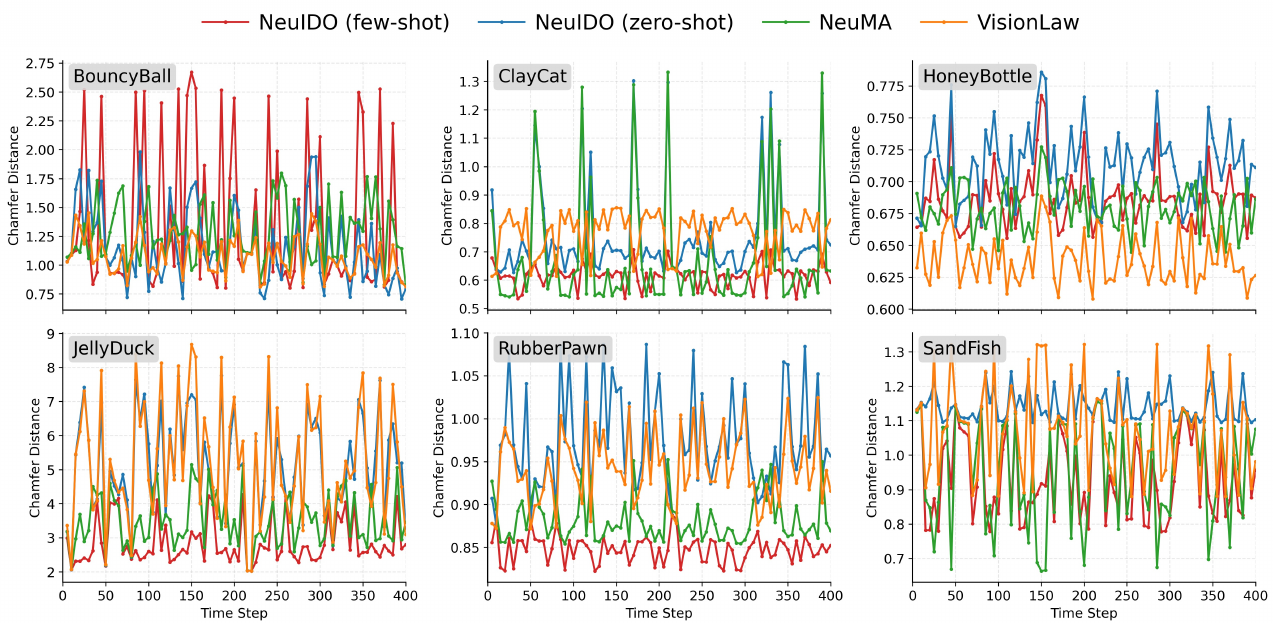}
	\caption{\textbf{Comparison of Chamfer Distance across different time steps on the synthetic dataset.} {\em NeuIDO} achieves competitive performance under the zero-shot setting and further improves after few-shot adaptation, obtaining the best overall accuracy and stability in most scenarios.}
    \label{fig:chamfer distance}
\end{figure*}

\section{More Experimental Results}
\subsection{Quantitative Comparison of Chamfer Distance}
To provide a finer-grained analysis of intrinsic dynamics modeling, we compare the chamfer distance of different methods across time steps and scenes, as shown in Fig.~\ref{fig:chamfer distance}. Unlike reporting only averaged chamfer distance metrics, this evaluation reveals whether a model can consistently track object evolution throughout the entire dynamic process.
Overall, {\em NeuIDO} exhibits lower and more stable error curves in most scenarios. Under the zero-shot setting, {\em NeuIDO} already achieves performance comparable to existing state-of-the-art methods, indicating its ability to learn a unified representation of intrinsic dynamics across multiple scenes.
After few-shot adaptation, the error is further reduced, leading to the best performance in the majority of cases.
Moreover, in scenarios with stronger temporal fluctuations, {\em NeuIDO} effectively suppresses error peaks and maintains more stable long-horizon predictions. 
These results demonstrate that {\em NeuIDO} achieves both high prediction accuracy and temporal stability while learning a unified representation of intrinsic dynamics.

\begin{figure*}[!t]
	\centering
	\includegraphics[width=1\linewidth]{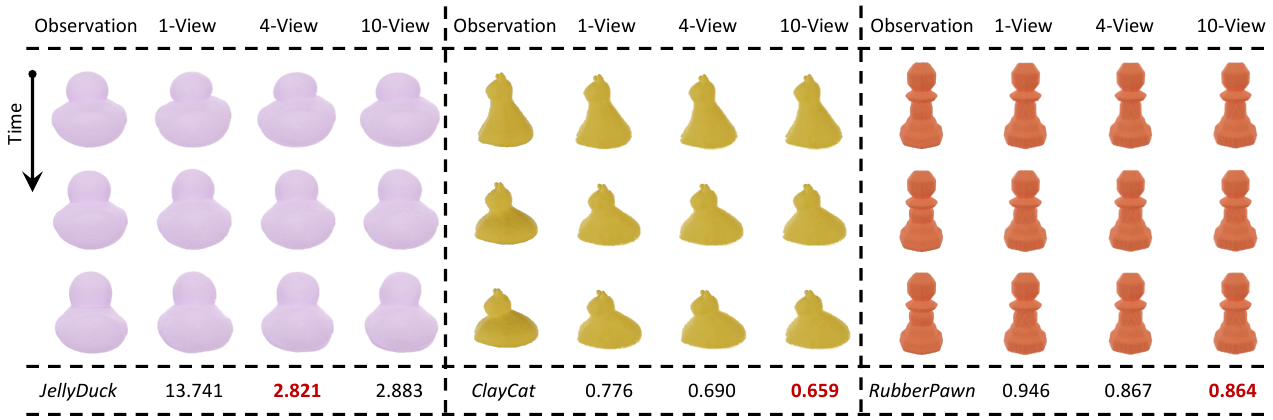}
	\caption{\textbf{Effect of the number of supervision views in few-shot adaptation}.
    Comparisons using 1, 4, and 10 input views during few-shot adaptation. Chamfer distances are shown in the bottom row. 
    Increasing the number of views improves adaptation quality by alleviating geometric ambiguity, while the improvement becomes marginal once sufficient coverage is reached.}
    \label{fig:diff view ablation few-shot apdation}
\end{figure*}

\begin{table*}[t]
\centering
\setlength{\tabcolsep}{12pt} 
{
\begin{tabular}{ccccc}
\hline
\textbf{View} & \textbf{Bun} & \textbf{Burger} & \textbf{Dog} & \textbf{Pig} \\ \hline
View-0 & 33.777 & 34.140 & 30.872 & 33.853 \\ 
View-1 & 32.916 & 33.144 & 30.806 & 32.627 \\
View-2 & 32.703 & 34.244 & 31.678 & 34.048\\ 
All & 34.944 & 34.186 & 34.959 & 35.295\\ 
\hline
\end{tabular}
}
\caption{\textbf{Monocular-video supervision on the real-world dataset.}}
\label{tab:Monocular Video Analysis}
\end{table*}

\subsection{Effect of Multi-view Supervision in Few-shot Adaptation}
We study the effect of the number of supervision views in the few-shot adaptation stage in Fig.~\ref{fig:diff view ablation few-shot apdation}. The trend closely matches the multi-view ablation in Stage II reported in the main paper.
Single-view supervision performs the worst across all scenes, due to incomplete observations and geometric ambiguity. Increasing the number of views substantially improves the results. In contrast, the gain from 4 to 10 views is limited. This suggests that beyond a certain point, additional views do not provide sufficiently richer supervision to further improve adaptation.
One possible reason is that visual supervision is intrinsically indirect: the dynamics of a 3D object must first be rendered onto the 2D image plane, and the optimization signal is then derived from image-space reconstruction errors. This projection from dynamic 3D states to 2D observations inevitably discards substantial geometric and temporal information, which limits the additional benefit of increasing the number of views beyond a certain point.
This observation motivates future efforts toward converting visual observations into denser and more direct signals for dynamics learning. Balancing performance and computational cost, we use 2 views as the default setting. 
We further evaluate {\em NeuIDO} under monocular-video supervision on the real-world dataset, using each of the three views independently.
As shown in Tab.~\ref{tab:Monocular Video Analysis}, monocular supervision generally lowers performance, with failures in challenging cases such as 'Dog', as a single view provides limited 3D deformation cues and makes dynamics identification ambiguous. 
Scenes such as “Pig” remain close to multi-view performance when dominant deformations are visible in the input view.

\begin{table*}[t]
\centering
\setlength{\tabcolsep}{12pt} 
{
\begin{tabular}{ccccc}
\hline
\textbf{Metric} & \textbf{Ball} & \textbf{Cat} & \textbf{Pawn} & \textbf{Fish} \\ \hline
Min & 1.235 & 0.607 & 0.848 & 0.923 \\ 
Max & 1.304 & 0.626 & 0.862 & 0.961 \\
Mean & 1.277 & 0.616 & 0.856 & 0.941\\ 
Variance & 5.228e-2 & 7.496e-3 & 5.033e-3 & 1.229e-2 \\ \hline
\end{tabular}
}
\caption{\textbf{Few-shot adaptation stability analysis.} We report the minimum, maximum, mean, and variance of chamfer distance over five random seeds for each scene.}
\label{tab:stability_analysis}
\end{table*}

\subsection{Few-shot Adaptation Stability Analysis}
We further evaluate the stability of few-shot adaptation by running {\em NeuIDO} with five random seeds. 
As shown in Tab.~\ref{tab:stability_analysis}, the adapted results exhibit small variations across seeds on all scenes, indicating that the few-shot adaptation process is stable and not sensitive to random initialization. 
This suggests that the pretrained dynamics representation provides a reliable prior for efficient scene-specific adaptation.

\begin{figure*}[!t]
	\centering
	\includegraphics[width=0.75\linewidth]{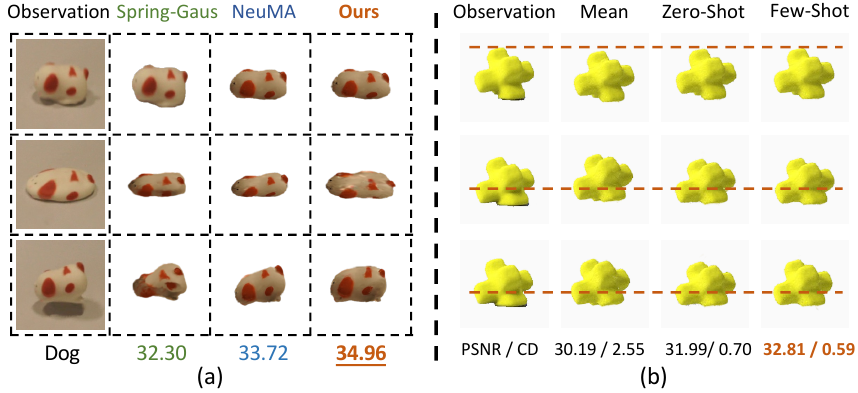}
	\caption{\textbf{Broader Evaluation.} (a) Additional real-world result for observed intrinsic dynamics modeling. (b) Additional unseen-scene generalization results.}
    \label{fig:broader_evaluation}
\end{figure*}

\subsection{Broader Evaluation}
We further provide broader evaluation results. 
Specifically, we add one real-world scene for observed intrinsic dynamics modeling (Sec.~\ref{sec: 4.2}) and five additional scenes for unseen-scenario generalization (Sec.~\ref{sec: 4.3.1}).
As shown in Fig.~\ref{fig:broader_evaluation} (a), {\em NeuIDO} achieves better reconstruction quality than Spring-Gaus and NeuMA on the additional real-world scene, further validating its ability to model observed intrinsic dynamics. 
Fig.~\ref{fig:broader_evaluation} (b) shows that {\em NeuIDO} also generalizes well to unseen scenes: zero-shot inference already produces dynamics closer to the observations than the mean baseline, and few-shot adaptation further improves both PSNR and chamfer distance. 
These results are consistent with the trends reported in Sec.~\ref{sec: 4.2} and Sec.~\ref{sec: 4.3.1}.

\section{Physics-Informed 4D Generation Pipeline}

Physics-informed 4D generation aims to model dynamic scenes by coupling physical simulation with differentiable visual representations. 
In this framework, material dynamics are governed by continuum mechanics, simulated using a numerical solver, and rendered through a differentiable scene representation that enables supervision from visual observations.

\subsection{Continuum Mechanics and Constitutive Laws}

In continuum mechanics~\cite{continuum}, the motion and deformation of materials are governed by the conservation of mass and momentum:
\begin{equation}
\frac{D\rho}{Dt} + \rho \nabla \cdot \mathbf{v} = 0, \quad
\rho \frac{D\mathbf{v}}{Dt} = \nabla \cdot \mathbf{P} + \rho \mathbf{g},
\end{equation}
where $\rho$ denotes the density, $\mathbf{v}$ the velocity field, $\mathbf{g}$ the gravitational acceleration, and $\mathbf{P}$ the stress tensor. 
The stress tensor is determined by the material constitutive law, which describes how a material responds to deformation. 
Therefore, the governing system becomes solvable only after specifying a constitutive relation for $\mathbf{P}$.

Within the MPM framework, two types of constitutive laws are typically required: 
(i) an elastic constitutive law that models reversible elastic responses, and 
(ii) a plastic constitutive law that captures irreversible plastic deformation. 
These relations can be written as
\begin{equation}
\varphi_E(\mathbf{F}; \theta_E) \mapsto \boldsymbol{\tau}, \quad
\varphi_P(\mathbf{F}; \theta_P) \mapsto \mathbf{F}^{\text{corrected}},
\end{equation}
where $\varphi_E$ and $\varphi_P$ denote the elastic and plastic constitutive laws, respectively. $\mathbf{F}$ denotes the deformation gradient, $\boldsymbol{\tau}$ is the Kirchhoff stress tensor, and $\mathbf{F}^{\text{corrected}}$ represents the deformation gradient after plastic return mapping. 
The parameters $\theta_E$ and $\theta_P$ represent material-specific properties. 
Although many classical constitutive laws have been proposed, they often struggle to capture the diversity and nonlinear behaviors of complex real-world materials.

\subsection{Material Point Method}
\label{appendix:MPM}

To simulate the above continuum equations, we adopt the Material Point Method (MPM)~\cite{MPM}, a hybrid simulation approach that combines Lagrangian particles with an Eulerian grid. 
MPM discretizes the material into particles that carry physical quantities such as mass, velocity, and deformation gradients, while grid nodes are used to compute intermediate physical interactions.
Each particle $p$ stores a set of physical attributes, including mass $m$, density $\rho$, volume $V$, velocity $\mathbf{v}$, deformation gradient $\mathbf{F}$, and velocity gradient $\mathbf{C}$. 
At each simulation step, MPM performs three stages: particle-to-grid (P2G) transfer, grid update, and grid-to-particle (G2P) transfer.
During the P2G stage, particle momentum and mass are transferred to grid nodes:
\begin{align}
m_i^{t+1} &= \sum_p w_{ip} m_p, \\
(m\mathbf{v})_i^{t+1} &= \sum_p w_{ip} 
\left[m_p \mathbf{v}_p^t + m_p \mathbf{C}_p^t(\mathbf{x}_i-\mathbf{x}_p^t)\right],
\end{align}
where $w_{ip}$ is the B-spline kernel weight between particle $p$ and grid node $i$.
Grid nodes then update their velocities by solving the momentum equation:
\begin{align}
\mathbf{v}_i^t &= (m\mathbf{v}_i)^t / m_i^t, \\
\mathbf{f}_{i,in}^t &= -\sum_p \boldsymbol{\tau}_p^t \nabla w_{ip} V_p, \\
\mathbf{v}_i^{t+1} &= \mathbf{v}_i^t + \Delta t 
\left(\mathbf{f}_{i,in}/m_i + \mathbf{g}\right).
\end{align}
Finally, updated grid quantities are interpolated back to particles:
\begin{align}
\mathbf{v}_p^{t+1} &= \sum_i w_{ip} \mathbf{v}_i^{t+1}, \\
\mathbf{x}_p^{t+1} &= \mathbf{x}_p^t + \Delta t \mathbf{v}_p^{t+1}, \\
\mathbf{C}_p^{t+1} &= \frac{4}{\Delta x^2} \sum_i w_{ip} 
\mathbf{v}_i^{t+1}(\mathbf{x}_i-\mathbf{x}_p^t)^T.
\end{align}
The deformation gradient is then updated as
\begin{align}
\mathbf{F}_p^{tr} &= (\mathbf{I}+\Delta t \mathbf{C}_p^{t+1})\mathbf{F}_p^t, \\
\mathbf{F}_p^{t+1} &= \varphi_P(\mathbf{F}_p^{tr}), \\
\boldsymbol{\tau}_p^{t+1} &= \varphi_E(\mathbf{F}_p^{t+1}),
\end{align}
where $\varphi_E$ and $\varphi_P$ denote the elastic and plastic constitutive laws, respectively. 
Through these steps, MPM advances the physical state of the system over time.

\subsection{Physics-Integrated 3D Gaussian Representation}

To connect physical simulation with visual observations, we adopt the 3D Gaussian Splatting (3DGS) representation~\cite{3DGS}. 
3DGS represents a scene using a set of anisotropic Gaussian kernels
\[
\mathcal{S} = \{\mathbf{x}_i,\mathbf{A}_i,\alpha_i,\mathbf{\mathcal{C}}_i\}_{i\in\mathcal{K}},
\]
where $\mathbf{x}_i$, $\mathbf{A}_i$, $\alpha_i$, and $\mathbf{\mathcal{C}}_i$ denote the center position, covariance matrix, opacity, and spherical harmonic coefficients of each Gaussian kernel.
Given a camera view, the color of a pixel is computed through alpha compositing:
\begin{equation}
\label{eq:render}
\mathbf{C} = 
\sum_{i\in\mathcal{N}}
\sigma_i \mathbf{SH}(d_i,\mathcal{C}_i)
\prod_{j=1}^{i-1}(1-\sigma_j),
\end{equation}
where $\sigma_i$ denotes the effective opacity and $\mathbf{SH}$ computes view-dependent color using spherical harmonics.
Unlike implicit neural representations such as NeRF, 3DGS provides an explicit and Lagrangian representation that naturally aligns with particle-based simulation. 
Following PhysGaussian~\cite{Physgaussian}, Gaussian kernels are treated as physical particles, allowing their motion and deformation to be simulated using MPM.
Specifically, given simulation conditions and constitutive laws, the MPM simulator predicts particle displacement and deformation:
\begin{align}
\mathbf{x}^{t+1},\mathbf{F}^{t+1} &= \mathbf{\Phi}(\mathcal{S}^t), \\
\mathbf{A}^{t+1} = \mathbf{F}&^{t+1}\mathbf{A}^t(\mathbf{F}^{t+1})^T,
\end{align}
where $\mathbf{\Phi}$ denotes the differentiable MPM simulator. 
The updated deformation gradient $\mathbf{F}^{t+1}$ also transforms the Gaussian covariance $\mathbf{A}$, approximating the local deformation of the kernel.
After the simulation step, the updated Gaussian parameters define a 4D Gaussian representation of the dynamic scene.

\subsection{Visual Supervision via Differentiable Rendering}
The simulated 4D Gaussian representation can be rendered into images using Eq.~\ref{eq:render}. 
These rendered frames can then be compared with visual observations to compute reconstruction losses. 
Because both the MPM simulation and the rendering process are differentiable, gradients can be propagated back through the simulation pipeline, enabling optimization of learnable parameters within the physical system.
Through this pipeline, constitutive laws determine intrinsic dynamics, MPM simulates these dynamics, and differentiable rendering provides visual supervision, enabling learning of intrinsic dynamics directly from visual observations.

\section{Limitation and Future Work}
Despite the encouraging results, {\em NeuIDO} still has several limitations.
First, the training in Stage II relies solely on visual supervision. 
Our multi-view ablation experiments show that visual observations provide only limited supervisory signals for intrinsic dynamics. 
Because the dynamics model is optimized indirectly through a rendering-based reconstruction pipeline, the supervision becomes weak and ambiguous. 
Moreover, increasing the number of input views brings only marginal improvements, indicating that visual supervision alone is insufficient for reliably constraining intrinsic dynamics.
Second, the proposed neural dynamics operator is fully parameterized by neural networks. 
Although this design provides strong representational capacity, the lack of explicit physical constraints may lead the model to overfit visual observations rather than learn the true underlying dynamics, similar to issues observed in prior neural constitutive modeling approaches such as NeuMA.
In future work, we plan to address these limitations by extracting richer and more informative supervisory signals from visual observations and introducing more explicit physical constraints into the operator formulation. 
We believe such improvements will further advance the development of physics-informed 4D generation and contribute toward more reliable world models.

\section{Approximation of the Neural Intrinsic Dynamics Operator}


\subsection{Overall.}
Having parameterized the neural intrinsic dynamics operator $\widehat{\mathcal{G}}$, we now establish its theoretical capability to approximate the ground-truth continuous mapping $\mathcal{G}:\mathcal{V}\rightarrow\mathcal{C}(K_{C})$. Due to the finite-dimensional compression of the video encoder $\mathbf{E}$, we assume there exists a fundamental representation error limit bounded by $\epsilon_{enc} \ge 0$. Building upon the universal approximation theorem for operator networks, we show that our branch-trunk architecture can universally approximate $\mathcal{G}$ up to this limit.

\textbf{Theorem 1 (Approximation theorem for Neural Intrinsic Dynamics Operator).} \textit{Let $\sigma$ be a continuous nonpolynomial activation function. Under the assumptions above, for any desired approximation error $\epsilon > 0$, there exist positive integers $n, D$ and parameters $\mathbf{a}_{ki} \in \mathbb{R}^d$, $\mathbf{w}_k \in \mathbb{R}^{d_c}$, $c_{ki}, \beta_{ki}, \zeta_k \in \mathbb{R}$ (for $i=1,...,n$ and $k=1,...,D$) such that the branch-trunk operator $\widehat{\mathcal{G}}$ satisfies:}
\begin{equation}
\left| \mathcal{G}(\mathbf{u})(\boldsymbol{\xi}) - \sum_{k=1}^D \underbrace{\left( \sum_{i=1}^n c_{ki}\sigma(\mathbf{a}_{ki}^\top \mathbf{E}(\mathbf{u}) + \beta_{ki}) \right)}_{\textnormal{branch}} \underbrace{\sigma(\mathbf{w}_k \cdot \boldsymbol{\xi} + \zeta_k)}_{\textnormal{trunk}} \right| < \epsilon_{enc} + \epsilon
\end{equation}
\textit{This inequality holds for all valid video functions $\mathbf{u} \in \mathcal{V}$ and constitutive inputs $\boldsymbol{\xi} \in K_C$. 
}

\subsection{Setting}

Let $K_V \subset \mathbb{R}^{d_v}$ be a compact video domain, and let $C(K_V)$ denote the Banach space of continuous video functions equipped with the uniform norm $\|\cdot\|_\infty$. Following Sec.~3.1 in the main text, let $\mathcal V \subset C(K_V)$ be a compact subset representing the set of valid video observations of interest. Let $K_C \subset \mathbb{R}^{d_c}$ be the compact constitutive-input domain. The target world-dynamics operator is a continuous mapping $\mathcal G:\mathcal V \to C(K_C)$.

The neural intrinsic dynamics operator adopts a branch--trunk form. Given an input video function $\mathbf u \in \mathcal V$, the dynamics encoder $E:\mathcal V \to \mathbb{R}^d$ extracts a latent representation $E(\mathbf u)$, and the branch network maps it to coefficients, while the trunk network parameterizes basis functions over $\boldsymbol\xi \in K_C$.

\subsection{Assumptions}

We make the following assumptions.

\paragraph{(A1) Compactness.}
$\mathcal V \subset C(K_V)$ and $K_C \subset \mathbb{R}^{d_c}$ are compact.

\paragraph{(A2) Continuity of the target operator.}
$\mathcal G:\mathcal V \to C(K_C)$ is continuous.

\paragraph{(A3) Approximate latent factorization.}
There exists a continuous encoder $E:\mathcal V \to \mathbb{R}^d$ and a continuous map $\bar{\mathcal G}:E(\mathcal V)\to C(K_C)$ such that
\[
\left|\mathcal G(\mathbf u)(\boldsymbol\xi)-\bar{\mathcal G}(E(\mathbf u))(\boldsymbol\xi)\right|
\le \epsilon_{\mathrm{enc}},
\qquad \forall\, \mathbf u\in\mathcal V,\ \forall\, \boldsymbol\xi\in K_C,
\]
for some $\epsilon_{\mathrm{enc}} \ge 0$.

\paragraph{(A4) Universal approximation families.}
Let $\sigma$ be a continuous non-polynomial activation function. Then standard feed-forward neural networks with activation $\sigma$ are universal approximators on compact subsets of finite-dimensional Euclidean spaces.

\subsection{Auxiliary Lemmas}

\paragraph{Lemma 1.}
If $\mathcal V$ is compact and $E:\mathcal V\to\mathbb{R}^d$ is continuous, then $S:=E(\mathcal V)\subset \mathbb{R}^d$ is compact.

\paragraph{Proof.}
Since continuous images of compact sets are compact, $S=E(\mathcal V)$ is compact.
\qed

\paragraph{Lemma 2.}
Let $S\subset\mathbb{R}^d$ be compact, and let $\bar{\mathcal G}:S\to C(K_C)$ be continuous. Define $H:S\times K_C\to\mathbb{R}$ by $H(\mathbf s,\boldsymbol\xi):=\bar{\mathcal G}(\mathbf s)(\boldsymbol\xi)$. Then $H$ is continuous on $S\times K_C$.

\paragraph{Proof.}
Take any sequence $(\mathbf s_n,\boldsymbol\xi_n)\to(\mathbf s,\boldsymbol\xi)$ in $S\times K_C$. Then
\[
|H(\mathbf s_n,\boldsymbol\xi_n)-H(\mathbf s,\boldsymbol\xi)|
\le
|H(\mathbf s_n,\boldsymbol\xi_n)-H(\mathbf s_n,\boldsymbol\xi)|
+
|H(\mathbf s_n,\boldsymbol\xi)-H(\mathbf s,\boldsymbol\xi)|.
\]
For the second term, by continuity of $\bar{\mathcal G}$ from $S$ into $C(K_C)$,
\[
|H(\mathbf s_n,\boldsymbol\xi)-H(\mathbf s,\boldsymbol\xi)|
=
|\bar{\mathcal G}(\mathbf s_n)(\boldsymbol\xi)-\bar{\mathcal G}(\mathbf s)(\boldsymbol\xi)|
\le
\|\bar{\mathcal G}(\mathbf s_n)-\bar{\mathcal G}(\mathbf s)\|_\infty
\to 0.
\]
For the first term, since $\bar{\mathcal G}(S)\subset C(K_C)$ is compact, it is equicontinuous on $K_C$. Hence, as $\boldsymbol\xi_n\to\boldsymbol\xi$,
\[
|H(\mathbf s_n,\boldsymbol\xi_n)-H(\mathbf s_n,\boldsymbol\xi)|
=
|\bar{\mathcal G}(\mathbf s_n)(\boldsymbol\xi_n)-\bar{\mathcal G}(\mathbf s_n)(\boldsymbol\xi)|
\to 0.
\]
Therefore, $H$ is continuous on $S\times K_C$.
\qed

\paragraph{Lemma 3.}
Let $S\subset\mathbb{R}^d$ and $K_C\subset\mathbb{R}^{d_c}$ be compact, and let $H:S\times K_C\to\mathbb{R}$ be continuous. Then for any $\varepsilon>0$, there exist a positive integer $D$ and continuous functions $\beta_k:S\to\mathbb{R}$ and $\phi_k:K_C\to\mathbb{R}$ for $k=1,\dots,D$ such that
\[
\left|
H(\mathbf s,\boldsymbol\xi)-\sum_{k=1}^D \beta_k(\mathbf s)\phi_k(\boldsymbol\xi)
\right|
<\varepsilon,
\qquad \forall\, \mathbf s\in S,\ \forall\, \boldsymbol\xi\in K_C.
\]

\paragraph{Proof sketch.}
This follows from the Stone--Weierstrass theorem on compact product spaces: finite sums of separable functions of the form $\beta(\mathbf s)\phi(\boldsymbol\xi)$ are dense in $C(S\times K_C)$.
\qed

\subsection{Proof of Theorem~1}

\paragraph{Theorem 1.}
Let $\sigma$ be a continuous non-polynomial activation function. Under Assumptions (A1)--(A4), for any $\epsilon>0$, there exist positive integers $n,D$ and parameters
\[
\mathbf a_{ki}\in\mathbb R^d,\quad
\mathbf w_k\in\mathbb R^{d_c},\quad
c_{ki},\beta_{ki},\zeta_k\in\mathbb R,
\qquad i=1,\dots,n,\;k=1,\dots,D,
\]
such that, for all $\mathbf u\in\mathcal V$ and $\boldsymbol\xi\in K_C$,
\[
\left|
\mathcal G(\mathbf u)(\boldsymbol\xi)-
\sum_{k=1}^{D}
\underbrace{\left(\sum_{i=1}^{n} c_{ki}\sigma(\mathbf a_{ki}^{\top}E(\mathbf u)+\beta_{ki})\right)}_{\textnormal{branch}}
\underbrace{\sigma(\mathbf w_k \cdot \boldsymbol\xi+\zeta_k)}_{\textnormal{trunk}}
\right|
<\epsilon_{\mathrm{enc}}+\epsilon.
\]

\paragraph{Proof.}

By Assumption~(A3), for all $\mathbf u\in\mathcal V$ and $\boldsymbol\xi\in K_C$,
\[
\left|\mathcal G(\mathbf u)(\boldsymbol\xi)-\bar{\mathcal G}(E(\mathbf u))(\boldsymbol\xi)\right|
\le \epsilon_{\mathrm{enc}}.
\]
Let $S:=E(\mathcal V)$. By Lemma~1, $S$ is compact. Define $H(\mathbf s,\boldsymbol\xi):=\bar{\mathcal G}(\mathbf s)(\boldsymbol\xi)$ for $\mathbf s\in S$ and $\boldsymbol\xi\in K_C$. By Lemma~2, $H$ is continuous on $S\times K_C$. Fix an arbitrary $\epsilon>0$.

\medskip
\noindent\textbf{Step 1 (Separable approximation).}
By Lemma~3, there exist a positive integer $D$ and continuous functions $\{\beta_k\}_{k=1}^D$ on $S$ and $\{\phi_k\}_{k=1}^D$ on $K_C$ such that
\begin{equation}\label{eq:step1}
\left|
H(\mathbf s,\boldsymbol\xi)-\sum_{k=1}^D \beta_k(\mathbf s)\,\phi_k(\boldsymbol\xi)
\right|
<\frac{\epsilon}{3},
\qquad \forall\, \mathbf s\in S,\ \forall\, \boldsymbol\xi\in K_C.
\end{equation}

\medskip
\noindent\textbf{Step 2 (Trunk approximation and re-indexing).}
Set $M_\beta:=\max_{1\le k\le D}\|\beta_k\|_\infty$. If $M_\beta=0$ then $H$ is already within $\epsilon/3$ of zero and the theorem follows trivially; we therefore assume $M_\beta>0$. For each $k=1,\dots,D$, since $\phi_k$ is continuous on the compact set $K_C\subset\mathbb{R}^{d_c}$, Assumption~(A4) guarantees the existence of a one-hidden-layer network
\[
\tilde\phi_k(\boldsymbol\xi)
:=\sum_{j=1}^{J_k} d_{kj}\,\sigma(\mathbf w_{kj}\cdot\boldsymbol\xi+\zeta_{kj})
\]
satisfying $\|\phi_k-\tilde\phi_k\|_\infty<\delta_2$, where we choose
\[
\delta_2:=\frac{\epsilon}{3\,D\,M_\beta}.
\]
Then, for all $\mathbf s\in S$ and $\boldsymbol\xi\in K_C$,
\begin{equation}\label{eq:step2}
\left|
\sum_{k=1}^D \beta_k(\mathbf s)\,\phi_k(\boldsymbol\xi)
-\sum_{k=1}^D \beta_k(\mathbf s)\,\tilde\phi_k(\boldsymbol\xi)
\right|
\le \sum_{k=1}^D |\beta_k(\mathbf s)|\;\|\phi_k-\tilde\phi_k\|_\infty
< D\,M_\beta\,\delta_2
=\frac{\epsilon}{3}.
\end{equation}
We now expand the product and re-index. Set $D':=\sum_{k=1}^D J_k$ and enumerate the pairs $(k,j)$ as a single index $m=1,\dots,D'$. For each $m$, define
\[
\gamma_m(\mathbf s):=d_{kj}\,\beta_k(\mathbf s),
\qquad
\mathbf w_m:=\mathbf w_{kj},
\qquad
\zeta_m:=\zeta_{kj},
\]
where $(k,j)$ is the pair corresponding to $m$. This re-indexing is an exact algebraic identity:
\[
\sum_{k=1}^D \beta_k(\mathbf s)\,\tilde\phi_k(\boldsymbol\xi)
=\sum_{m=1}^{D'}\gamma_m(\mathbf s)\,\sigma(\mathbf w_m\cdot\boldsymbol\xi+\zeta_m).
\]

\medskip
\noindent\textbf{Step 3 (Branch approximation).}
Since $\sigma$ is continuous and $K_C$ is compact, each trunk unit is bounded on $K_C$. Set
\[
M_\sigma:=\max_{1\le m\le D'}\sup_{\boldsymbol\xi\in K_C}|\sigma(\mathbf w_m\cdot\boldsymbol\xi+\zeta_m)|.
\]
If $M_\sigma=0$ then the approximation is already within $2\epsilon/3$ and we are done; assume $M_\sigma>0$. Each $\gamma_m$ is continuous on the compact set $S\subset\mathbb{R}^d$, so by Assumption~(A4), for each $m=1,\dots,D'$, there exists a one-hidden-layer network of width $n_m$,
\[
b_m(\mathbf s):=\sum_{i=1}^{n_m} c_{mi}\,\sigma(\mathbf a_{mi}^{\top}\mathbf s+\beta_{mi}),
\]
satisfying $\|\gamma_m-b_m\|_\infty<\delta_1$, where we choose
\[
\delta_1:=\frac{\epsilon}{3\,D'\,M_\sigma}.
\]
To obtain a uniform width across all branch networks, set $n:=\max_{1\le m\le D'}n_m$ and, for each $m$ with $n_m<n$, pad with zero coefficients $c_{mi}=0$, $\mathbf a_{mi}=\mathbf 0$, $\beta_{mi}=0$ for $i=n_m+1,\dots,n$. This does not change the network output. Hence, for all $\mathbf s\in S$ and $\boldsymbol\xi\in K_C$,
\begin{equation}\label{eq:step3}
\begin{aligned}
&
\left|
\sum_{m=1}^{D'}\gamma_m(\mathbf s)\,\sigma(\mathbf w_m\cdot\boldsymbol\xi+\zeta_m)
-\sum_{m=1}^{D'} b_m(\mathbf s)\,\sigma(\mathbf w_m\cdot\boldsymbol\xi+\zeta_m)
\right|
\\
&\le
\sum_{m=1}^{D'} |\gamma_m(\mathbf s)-b_m(\mathbf s)|\;|\sigma(\mathbf w_m\cdot\boldsymbol\xi+\zeta_m)|
\\
&<
D'\,\delta_1\,M_\sigma
=
\frac{\epsilon}{3}.
\end{aligned}
\end{equation}

\medskip
\noindent\textbf{Combining the bounds.}
Applying the triangle inequality to \eqref{eq:step1}, \eqref{eq:step2}, and \eqref{eq:step3}, we obtain, for all $\mathbf s\in S$ and $\boldsymbol\xi\in K_C$,
\[
\left|
H(\mathbf s,\boldsymbol\xi)
-\sum_{m=1}^{D'}\left(\sum_{i=1}^{n} c_{mi}\,\sigma(\mathbf a_{mi}^{\top}\mathbf s+\beta_{mi})\right)
\sigma(\mathbf w_m\cdot\boldsymbol\xi+\zeta_m)
\right|
<\frac{\epsilon}{3}+\frac{\epsilon}{3}+\frac{\epsilon}{3}=\epsilon.
\]
Finally, substituting $\mathbf s=E(\mathbf u)$ and recalling $H(E(\mathbf u),\boldsymbol\xi)=\bar{\mathcal G}(E(\mathbf u))(\boldsymbol\xi)$, we obtain by the triangle inequality and Assumption~(A3):
\[
\left|
\mathcal G(\mathbf u)(\boldsymbol\xi)-
\sum_{m=1}^{D'}
\left(\sum_{i=1}^{n} c_{mi}\,\sigma(\mathbf a_{mi}^{\top}E(\mathbf u)+\beta_{mi})\right)
\sigma(\mathbf w_m\cdot\boldsymbol\xi+\zeta_m)
\right|
\]
\[
\begin{aligned}
&\le
\underbrace{\left|\mathcal G(\mathbf u)(\boldsymbol\xi)-\bar{\mathcal G}(E(\mathbf u))(\boldsymbol\xi)\right|}_{\le\;\epsilon_{\mathrm{enc}}}
+
\underbrace{\left|\bar{\mathcal G}(E(\mathbf u))(\boldsymbol\xi)-\sum_{m=1}^{D'} b_m(E(\mathbf u))\,\sigma(\mathbf w_m\cdot\boldsymbol\xi+\zeta_m)\right|}_{<\;\epsilon}
\\[4pt]
&<
\epsilon_{\mathrm{enc}}+\epsilon.
\end{aligned}
\]
for all $\mathbf u\in\mathcal V$ and $\boldsymbol\xi\in K_C$. Renaming $D'\to D$ completes the proof.
\qed

\end{document}